\documentclass{article}

\usepackage[a4paper, total={6in, 8in}]{geometry}
\usepackage[utf8]{inputenc} 
\usepackage[T1]{fontenc}    
\usepackage{hyperref}       
\usepackage{url}            
\usepackage{booktabs}       
\usepackage{amsfonts}       
\usepackage{nicefrac}       
\usepackage{microtype}      
\usepackage{xcolor}         

\usepackage{times}
\usepackage{soul}
\usepackage{url}
\usepackage[small]{caption}
\usepackage{graphicx}
\usepackage{amsmath}
\usepackage{amsthm}
\usepackage{amssymb}
\usepackage{bm}
\usepackage{booktabs}
\usepackage{algorithm}
\usepackage{algorithmic}
\usepackage{xcolor}
\usepackage{subfigure}
\usepackage{subcaption}

\usepackage{multirow}
\usepackage{diagbox}
\usepackage{graphicx}
\usepackage{booktabs}
\usepackage{hyperref}
\usepackage{url}
\usepackage[normalem]{ulem}
\usepackage[inline]{enumitem}
\usepackage{makecell}
\usepackage{soul}
\usepackage{footmisc}
\usepackage{tikz}

\usepackage{titlesec}
\usepackage{wrapfig}
\usepackage{pifont}
\usepackage{enumitem}

\titleformat{\subsection}[runin]{\normalfont\bfseries}{\thesubsection}{1em}{}[:]
\titlespacing*{\subsection}{0pt}{\baselineskip}{\baselineskip}

\titleformat{\subsubsection}[runin]{\normalfont\bfseries}{\thesubsubsection}{1em}{}[:]
\titlespacing*{\subsubsection}{0pt}{\baselineskip}{\baselineskip}

\title{\textbf{Tiny Time Mixers (TTMs): Fast Pre-trained Models for Enhanced Zero/Few-Shot Forecasting of Multivariate Time Series}}

\author{
  Vijay Ekambaram
  \and
  Arindam Jati 
  \and
  Pankaj Dayama 
  \and
  Sumanta Mukherjee 
  \and
  Nam H. Nguyen 
  \and
  Wesley M. Gifford 
  \and
  Chandra Reddy 
  \and
  Jayant Kalagnanam
}

\date{IBM Research\footnote{Corresponding e-mail: \texttt{vijaye12@in.ibm.com}.}}

\newcommand{\ie}{i.e., }

\newcommand{\etc}{\textit{etc.}}
\newcommand{\wrt}{\textit{w.r.t.}~}

\newcommand{\ttms}{TTM$_\textit{B}$}
\newcommand{\ttmm}{TTM$_\textit{E}$}
\newcommand{\ttml}{TTM$_\textit{A}$}
\newcommand{\ttmq}{TTM$_\textit{Q}$}

\newcommand{\moirais}{Moirai$_\textit{S}$}
\newcommand{\moiraib}{Moirai$_\textit{B}$}
\newcommand{\moirail}{Moirai$_\textit{L}$}

\newcommand{\chronust}{Chronos$_\textit{T}$}
\newcommand{\chronuss}{Chronos$_\textit{S}$}
\newcommand{\chronusb}{Chronos$_\textit{B}$}
\newcommand{\chronusl}{Chronos$_\textit{L}$}

\definecolor{v_blue}{RGB}{0, 50, 150} 

\begin{document}


\maketitle

\begin{abstract}

%
Large pre-trained models excel in zero/few-shot learning for language and vision tasks but face challenges in multivariate time series (TS) forecasting due to diverse data characteristics. Consequently, recent research efforts have focused on developing pre-trained TS forecasting models. These models, whether built from scratch or adapted from large language models (LLMs), excel in zero/few-shot forecasting tasks. However, they are limited by slow performance, high computational demands, and neglect of cross-channel and exogenous correlations. To address this, we introduce Tiny Time Mixers (TTM), a compact model (starting from 1M parameters) with effective transfer learning capabilities, trained exclusively on public TS datasets. TTM, based on the light-weight TSMixer architecture, incorporates innovations like adaptive patching, diverse resolution sampling, and resolution prefix tuning to handle pre-training on varied dataset resolutions with minimal model capacity. Additionally, it employs multi-level modeling to capture channel correlations and infuse exogenous signals during fine-tuning. TTM outperforms existing popular benchmarks in zero/few-shot forecasting  by (4-40\%), while reducing computational requirements significantly. Moreover, TTMs are lightweight and can be executed even on CPU-only machines, enhancing usability and fostering wider adoption in resource-constrained environments. The model weights for reproducibility and research use are available \href{https://huggingface.co/ibm/ttm-research-r2/}{\textcolor{blue}{here}}, while enterprise-use weights under the Apache license can be accessed as follows: the initial \ttmq variant \href{https://huggingface.co/ibm-granite/granite-timeseries-ttm-r1}{\textcolor{blue}{here}}, and the latest variants (\ttms, \ttmm, \ttml) weights are available \href{https://huggingface.co/ibm-granite/granite-timeseries-ttm-r2}{\textcolor{blue}{here}} (preferred use). The source code for the TTM model along with the usage scripts are available \href{https://github.com/ibm-granite/granite-tsfm/tree/main/tsfm_public/models/tinytimemixer}{\textcolor{blue}{here}}.\\ \\
\textit{[Accepted at the 38th Conference on Neural Information Processing Systems (NeurIPS 2024). This work is submitted to arXiv on Jan 08, 2024.] }

\end{abstract}

\section{Introduction}\label{sec:Introduction}

Multivariate time-series~(TS) forecasting entails predicting future values for multiple interrelated time series based on their historical values.
The channels\footnote{``Channel'' refers to an individual dimension in multivariate data (i.e.,\ multivariate or multichannel).} being forecast are called target variables, while those influencing the forecasts are termed exogenous variables. 
This field has seen significant advancements through the application of statistical and machine learning (ML) methods across various domains such as weather, traffic, retail, and energy.



\textbf{Related Work:} Recent advances in multivariate forecasting have been marked by the advent of Transformer-based~\cite{transformer} approaches, exemplified by models like PatchTST~\cite{patchtst}, Autoformer~\cite{autoformer}, and FEDFormer~\cite{fedformer}. These models have demonstrated notable improvements over traditional statistical and ML methods. Furthermore, architectures based on MLP-Mixer~\cite{mlpmixer}, such as TSMixer~\cite{tsmixer} and TimeMixer~\cite{timemixer}, have emerged as efficient Transformer alternatives, boasting 2-3X reduced compute requirements with no accuracy compromise compared to their Transformer counterparts.

\begin{wrapfigure}{l}{0.45\textwidth}
\vspace{-0.7cm}
\includegraphics[width=\linewidth]{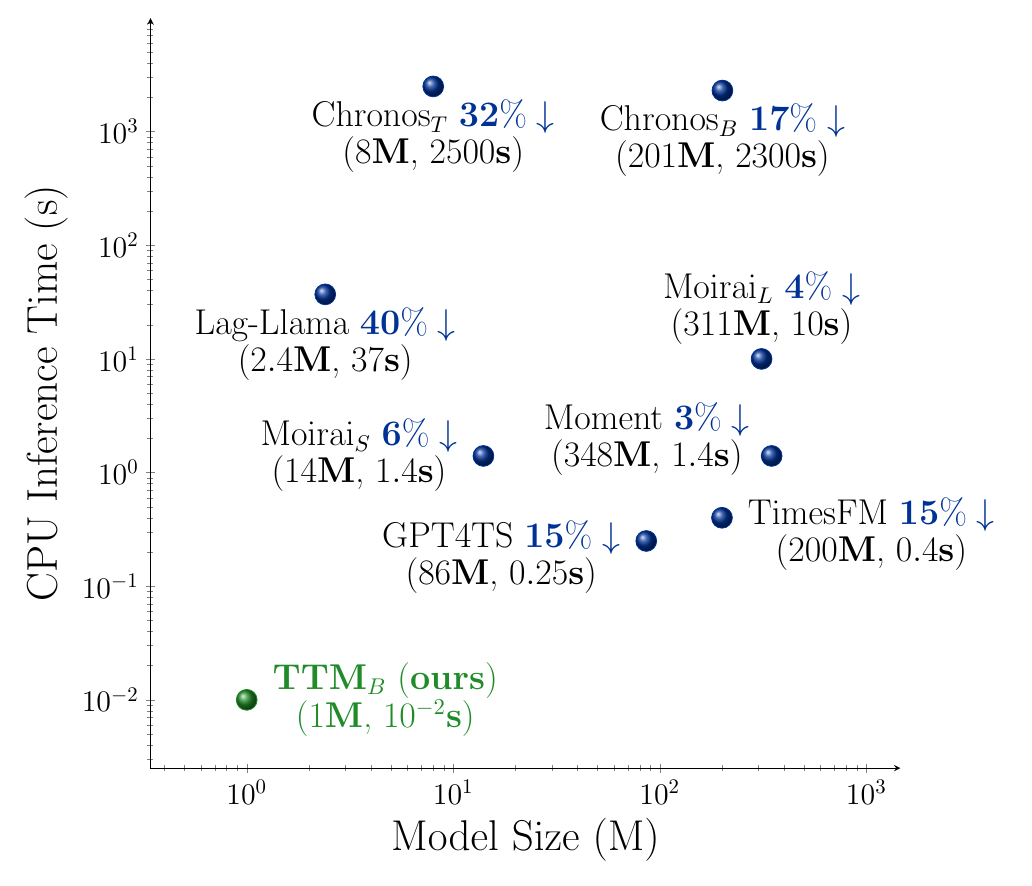}
\caption{\textbf{Size, time, and accuracy overview} of \ttms~ vs.\ open-sourced pre-trained TS benchmarks\protect\footnotemark.  We plot each model based on its model size and per batch CPU inference time. The \textcolor{v_blue}{X\%} mentioned for each baseline indicates that the baseline's forecast is \textcolor{v_blue}{X\%} less accurate compared to TTM's forecast in the evaluation benchmarks. Full details in Tables [\ref{tab:n_zs_moirai_avg}--\ref{tab:n_hp_avg}]. }
\label{fig:ttm_q_overview}
\vspace{-0.3cm}

\end{wrapfigure}
Recently, there has been substantial interest in the research community to build general pre-trained or foundation models~(FMs) for TS forecasting that can successfully transfer the learning to unseen target TS dataset, similar to the successes in NLP and vision tasks. 
However, pre-training in the TS domain is particularly challenging due to the limited public availability of the datasets and the diverse nature across applications. 
Early in 2024, this interest culminated in the release of several ``large'' and ``massive'' TS pre-trained models for forecasting, generating considerable excitement in the research community. Among these releases were 
Moment\cite{moment}\footnote{\label{cwork}Work done concurrently with this research.}, TimesFM~\cite{timesfm}\textsuperscript{\ref{cwork}}, Chronos\cite{chronos}\textsuperscript{\ref{cwork}}, Moirai\cite{moirai}\textsuperscript{\ref{cwork}}, and Lag-llama\cite{lagllama}\textsuperscript{\ref{cwork}} that successfully established strong benchmarks in zero-shot forecasting. In addition, there has been a trend towards leveraging pre-trained large language models (LLMs) for TS forecasting, treating the forecasting task as a form of cross-domain transfer learning. These universal cross-transfer approaches, exemplified by recent works such as LLMTime~\cite{llmtime}, Time-LLM~\cite{timellm}, and GPT4TS~\cite{gpt4ts}, exhibit promising outcomes in zero/few-shot forecasting scenarios. However, most of these ``large'' TS pre-trained models demand extremely high computational resources, given their scale ranges from several hundred million to billions of parameters. Given the recent surge in popularity of ``small'' language models\cite{phi}\cite{slm}\cite{tinygpt} that address practical resource and cost constraints in real-world industrial settings, this work considers the following question: \textit{Can ``tiny'' pre-trained models succeed in the TS domain too? If so, can they outperform the zero/few-shot forecasting results of ``large'' TS pre-trained models demanding significant computational resources and runtime?} Surprisingly, as we demonstrate in this work, the answer is \textit{yes}.

\footnotetext{Time-LLM and LLMTime are excluded here as we couldn't run them on CPUs, but their accuracy is compared later in experiments.}

Toward this, we propose \textbf{Multi-level Tiny Time Mixers~(TTM)}, a significantly smaller pre-trained model (starting from 1 million (M) parameters) for effective zero/few-shot multivariate forecasting. In particular, TTM supports channel correlations and exogenous signals, which are critical and practical business requirements in the context of multivariate forecasting, features lacking in many existing TS pretrained models. TTM is based on the light-weight TSMixer architecture that uses MLPMixer blocks interleaved with simple gated attention as alternatives to the quadratic time-consuming self-attention blocks in Transformers, which makes TTM pre-training and fine-tuning extremely fast. TTM is pre-trained using multiple public datasets ($\sim$1 billion (B) samples)  from the Monash and LibCity data repositories. Note that the datasets exhibit considerable diversity in  characteristics, such as different domains, temporal resolutions\footnote{Resolution refers to the sampling rate of the input time series (e.g., hourly, 10 minutes, 15 minutes, \etc)} (ranging from seconds to days), lengths, and numbers of channels. Pre-training on such heterogeneous datasets using extremely small models requires specialized architectural advancements. Hence, TTM proposes the following enhancements to the TSMixer architecture for resource-constrained pre-training/fine-tuning: (i) \textbf{adaptive patching (AP)} across layers, considering the varied suitability of patch lengths for different datasets, (ii) \textbf{diverse resolution sampling (DRS)}  
to augment the data for increasing coverage across different resolutions, (iii) \textbf{resolution prefix tuning (RPT)} to explicitly embed resolution information in the first patch, facilitating resolution-conditioned modeling while training on diverse datasets. 
Moreover, our approach leverages \textbf{multi-level modeling}, where TTMs are first pre-trained in a channel-independent way, and then fine-tuned with channel mixing to incorporate correlations across targets and exogenous channels in the target domain.

\textbf{Outline of TTM's key capabilities:}
\begin{enumerate*}[label=(\arabic*)]
    \item Amidst the prevalence of ``large'' pre-trained models demanding significant compute and training time, our work is the first to demonstrate the power of transfer learning using ``tiny'' TS pre-trained models for zero/few-shot forecasting. 
    \item Pre-training tiny models on heterogeneous multi-resolution datasets with extremely limited model capacity is challenging. Towards this, we propose various \textbf{architectural and training enhancements}, such as \textbf{AP}, \textbf{DRS}, and \textbf{RPT} for robust and resource-constrained pre-training/fine-tuning workflows (as defined above).
    \item TTM employs a \textbf{multi-level modeling strategy} to explicitly model channel correlations, and incorporate exogenous signals -- a crucial capability lacking in most of the existing pre-trained models.
    \item Through extensive evaluation of zero/few-shot forecasting on 11 datasets, we establish that TTM models, with sizes as small as 1M parameters, consistently outperform the forecasts of ``large'' TS pretrained models while offering significant computational benefits. Figure~\ref{fig:ttm_q_overview} highlights that TTM outperforms popular benchmarks in all three primary dimensions: size, runtime, and accuracy.
    \item Given their compact size, zero-shot inference and fine-tuning of TTM models can be easily executed with just one GPU or in CPU-only environments.
    This greatly enhances the  practical adoption and extended reach of our pre-trained models with ease of use. 
\end{enumerate*}

\begin{figure*}[h]
\vspace{-0.2cm}
  \begin{minipage}{0.95\textwidth}
     \centering
    \includegraphics[width=1\linewidth]{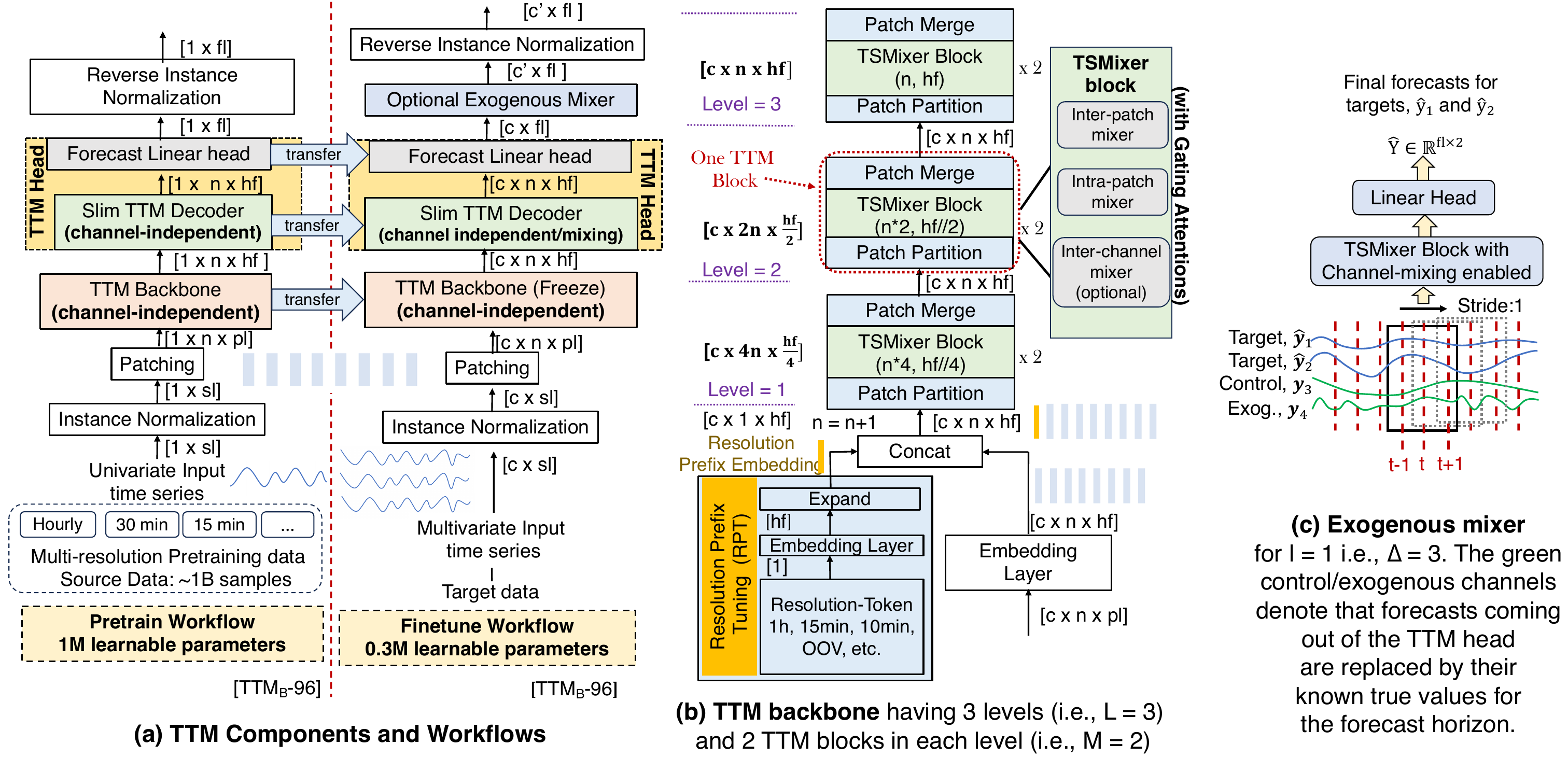}
    \caption{TTM overview (a) Refer to Sections~\ref{sec:ttm_components} and \ref{sec:ttm_workflows}, (b) Refer to Section~\ref{sec:Pre-training Workflow}, (c) Refer to Section~\ref{exog_section}}
    \label{fig:ttm_overview}
  \end{minipage}
  \vspace{-0.6cm}
\end{figure*}
\section{TTM Components}\label{sec:ttm_components}
Let $\bm{X} \in \mathbb{R}^{c \times sl}$ be a multivariate time series of length $sl$ and number of channels $c$. The forecasting task can be formally defined as predicting the future values $\bm{Y} \in \mathbb{R}^{c' \times fl}$ given the history/context $\bm{X}$. Here, $fl$ denotes the forecast length/horizon, and $c'$ denotes number of forecast channels, where $c' \le c$. 
The predictions from the model are denoted by $\hat{\bm{Y}} \in \mathbb{R}^{c' \times fl}$.
In a general multivariate forecasting task, each channel falls into one of the following categories:
\begin{enumerate*}[label=(\alph*)]
\item \textbf{Target variables (mandatory):} channels for which forecasts are required, 
\item \textbf{Exogenous variables (optional):} channels influencing the targets, with known or estimated values throughout the forecast horizon.
\end{enumerate*}

\subsection{Multi-level Modeling} \label{sec:Multi-level Modeling}
TTM follows a multi-level architecture consisting of four key components (see Figure~\ref{fig:ttm_overview}(a)):
\begin{enumerate*}[label=(\arabic*)]
    \item The \textbf{TTM backbone} is assembled using building blocks derived from the efficient TSMixer architecture~\cite{tsmixer}. TSMixer is based on MLP blocks interleaved with gated attention, that enable the mixing of features within patches, across patches and channels, surpassing existing Transformer-based TS approaches with minimal computational requirements. Since TSMixer was not designed to handle multi-resolution data with limited capacity, we introduce various novel enhancements to it as explained later.
    \item \textbf{TTM decoder} follows the same backbone architecture but is considerably smaller in size, approximately 10-20\% of the size of the backbone,
    \item \textbf{Forecast head} consists of a linear head designed to produce the forecast output, and 
    \item Optional \textbf{Exogenous mixer} serves to fuse exogenous data into the forecasting process. The TTM decoder and forecast head together constitute the \textbf{TTM head}, whose weights get updated during the fine-tuning process.
\end{enumerate*}
This multi-level model refactoring is required to dynamically change the working behavior of various components based on the workflow type, as explained in Section~\ref{sec:ttm_workflows}. In addition to the above primary components, there is also a preprocessing component as explained next.

\subsection{Preprocessing}\label{sec:Pre-processing}
As shown in Figure~\ref{fig:ttm_overview}(a) with colorless blocks, the historical time series $\bm{X}$ is first \textbf{normalized} per instance to have zero mean and unit standard deviation for each channel, to tackle any possible distribution shifts~\cite{patchtst,tsmixer}.
This process is reversed at the end before computing the loss.
The normalized data $\overline{\bm{X}}$ is subsequently \textbf{patched}, $\bm{X}_p \in \mathbb{R}^{c \times n \times pl}$, into $n$ non-overlapping windows, each of length $pl$, and then passed to the TTM backbone.
Patching, as introduced in \cite{patchtst}, has proven to be highly valuable for forecasting. Its effectiveness lies in preserving local semantic information, accommodating longer history, and reducing computation.

\section{TTM Methodology}\label{sec:ttm_workflows}
\subsection{Pre-training Workflow}\label{sec:Pre-training Workflow}
TTM operates in two stages: pre-training and fine-tuning (Figure~\ref{fig:ttm_overview}(a)). In the pre-training stage, we train the model on a large collection of diverse public datasets. Since the primary focus of TTM is forecasting, pre-training is modeled with a direct forecasting objective. 
TTM is first pre-trained in a univariate fashion with independent channels on all the existing datasets. 
Due to varied channel counts in pre-training datasets, modeling multivariate correlations is challenging here; it is addressed later during the fine-tuning stage. Multivariate pre-training datasets are initially transformed into independent univariate TS $(\bm{X}_1,\cdots,\bm{X}_N) \in \mathbb{R}^{c(=1) \times sl}$. 
These are pre-processed (Section~\ref{sec:Pre-processing}), and subsequently fed into the TTM backbone for multi-resolution pre-training. The output of the backbone $\bm{X}_h^L \in \mathbb{R}^{(c=1) \times n \times hf}$ is passed through the decoder and forecast head to produce the forecast $\hat{\bm{Y}} \in \mathbb{R}^{(c=1) \times fl }$ which is then reverse-normalized to return to the original scale. We pre-train the TTM with mean squared error (MSE) loss calculated over the forecast horizon: $\mathcal{L} = || \bm{Y} - \hat{\bm{Y}} ||_2^2$. 
Thus for a given input context length $sl$ and forecast length $fl$, we get a  pre-trained model capturing the common temporal forecasting dynamics and seasonal patterns as observed in the overall pre-training data. 

\subsubsection{Multi-Resolution Pre-training via TTM Backbone}\label{sec:TTM Backbone}
In TTM, our goal is to create models that are extremely small yet capable of generalizing well across a wide range of diverse datasets with varying resolutions. This is a significant challenge because the models can easily under-fit due to their small size. To tackle these challenges of resource-constrained pre-training, we introduce the following enhancements to the TSMixer backbone.




\textbf{Adaptive patching (AP):} The TTM backbone is crafted with an adaptive patching architecture where different layers of the backbone operate at varying patch lengths and numbers of patches. Since each dataset in the pre-training corpora may perform optimally at a specific patch length, this approach greatly aids in generalization when the pretraining datasets with different resolutions are introduced. Moreover, it helps in scenarios when the availability of the pre-training data is limited as adaptive patching quickly generalizes the model across different granularities.
As shown in Figure~\ref{fig:ttm_overview}(b), the patched data $\bm{X}_p\in \mathbb{R}^{c \times n \times pl}$ is passed through a embedding layer to project it to the patch hidden dimension, $\bm{X}_h \in \mathbb{R}^{c \times n \times hf}$.
If the resolution prefix tuning module is activated (as explained later), the resolution prefix is concatenated with $\bm{X}_h$. For notational simplicity, we denote the concatenated tensor with $\bm{X}_h$ as well. 
The TTM backbone consists of $L$ levels, each comprising $M$ TTM blocks with identical patch configurations.
The first block in the first level receives $\bm{X}_h$.
The first TTM block in the $i$-th level, $i=2,\ldots,L$, receives the processed data $\bm{X}_h^{(i-1)} \in \mathbb{R}^{c \times n \times hf}$ from the previous block.
Each TTM block is further comprised of a patch partition block, a vanilla TSMixer block, and a patch merging block. 
The patch partition block at level $i$ increases the number of patches by a factor of $K_i$ and reduces the patch dimension size by the same factor by reshaping $\bm{X}_h^{(i-1)} \in \mathbb{R}^{c \times n \times hf} \nonumber
\text{to } \bm{X}_h^i \in \mathbb{R}^{c \times (n \cdot K_i) \times (hf/K_i)}$, where $K_i = 2^{(L-i)}$. 
Figure~\ref{fig:ttm_overview}(b) shows the TTM backbone for $L=3$ and $M=2$.
Note that, we set $hf=m \cdot 2^{L-1}$ for some integer $m$.
Then, TSMixer is applied to the adapted data $\bm{X}_h^i$.
Finally, the output from TSMixer is again reshaped to its original shape (\ie $\mathbb{R}^{c \times n \times hf}$) in the patch merging block. In subsequent layers, for each increment in level $i$, the number of patches is halved and the patch dimension doubled.
This enables better generalization for small models as we pre-train across multiple datasets. The idea of adaptive patching is popular and very successful in the vision domain (e.g., Swin Transformers~\cite{swin-transformer}) and we successfully  apply it  to the TS domain to resolve multi-resolution issues in modelling diverse TS datasets.  Note that adaptive patching is enabled only in the backbone and not in the decoder, which is designed to be very lightweight.

\textbf{Augmentation via diverse resolution sampling (DRS):}
A significant challenge in TS pre-training datasets is the scarcity of public datasets with diverse resolutions. 
Generally, high-resolution datasets will account for a larger fraction of the samples given their finer sampling resolution. Without adjustment to the training strategy, this can lead to a model that is biased toward the finer resolution data. To overcome this,
different strategies are applied to high-resolution datasets to balance the volume of samples at lower resolutions and lead to more uniform coverage. Strategies used include: 1) averaging $k$ samples in sequential, non-overlapping windows to produce a lower resolution dataset; and 2) conventional decimation where only every $k$th sample is retained. In both cases, the integer $k$ is chosen to achieve the desired resolution from the resolution of the base dataset.
For example, from a 4-second resolution dataset, we derive multiple datasets at minutely ($k=15$) and hourly resolutions ($k=900$).
Note that the original high-resolution dataset remains within the pool of pre-training datasets. 
This methodology increases the number of datasets for each resolution which greatly improves the model performance.

\textbf{Resolution prefix tuning (RPT):} This technique explicitly learns and incorporates a new patch embedding as a learnable prefix into the input data based on the input resolution (see Figure~\ref{fig:ttm_overview}(b) and Table~\ref{tab:pre-train_datasets}). 
Similar to the concept of prefix tuning~\cite{prefix-tuning}, this approach provides an explicit signal to the model about the resolution for resolution-conditioned modeling. 
First, we map every resolution to a unique integer, which is then passed through an embedding layer to project it to the hidden dimension, $hf$. 
Subsequently, we expand the embedding across all channels to have a representation of shape $c\times 1\times hf$. 
This resolution-based learnable embedding is particularly beneficial in quickly modeling huge volumes of diverse resolution datasets with limited modelling capacity, as the model can easily decouple the data from different resolutions for resolution-conditioned modeling. In addition, RPT also helps in scenarios when the context length ($sl$) is short. In these scenarios, automatically detecting the resolution becomes a challenge for the model. Hence, by explicitly fusing the resolution information as a prefix, we can enhance the model's ability to learn effectively across resolutions without increasing its size.

\subsection{Fine-tuning Workflow}\label{sec:Fine-tuning Workflow}
In the fine-tuning workflow, we work with data from the \textit{target} domain that has no overlap with the pre-training datasets. We have three options here:
\begin{enumerate*}[label=(\alph*)]
    \item In \textbf{zero-shot} forecasting, we directly use the pre-trained model to evaluate on the \textit{test} part of the target data;
    \item In \textbf{few-shot} forecasting, we utilize only a tiny portion (5-10\%) of the \textit{train} part of the target data to quickly update the pre-trained weights of the \textit{TTM head}, and subsequently evaluate it on the \textit{test} part;
    \item In \textbf{full-shot} forecasting, we fine-tune the pre-trained weights of the \textit{TTM head} on the entire \textit{train} part of the target data, and then evaluate on the \textit{test} part.
\end{enumerate*}

The backbone is frozen during fine-tuning, and still operates in a channel-independent univariate fashion. 
However, the slim decoder in the TTM Head can be fine-tuned utilizing channel mixing or channel independence for multivariate or univariate target data, respectively.
If pure multivariate modeling is needed, then the channel-mixer block in all the TSMixer components (see Figure~\ref{fig:ttm_overview}(b)) in the decoder is enabled to explicitly capture the cross-channel correlations. The forecast head and reverse normalization perform similar operations as in the pre-training stage. 
The fine-tuning also optimizes the forecasting objective with MSE loss. 
This thoughtful multi-level design choice ensures that our backbone excels in channel-independent pre-training, enabling effective temporal correlation modeling across diverse datasets. Simultaneously, the decoder handles target-data-specific tasks like channel-correlation modeling and fine-tuning. In addition, if the target data has exogenous variables, then an exogenous mixer block is applied to the actual forecasts as explained next. 

\textbf{Exogenous Mixer Block:} 
\label{exog_section}
As described in Section~\ref{sec:ttm_components}, the future values of the exogenous channels are known in advance. Let the forecast from the forecast head be $\hat{\bm{Y}} \in \mathbb{R}^{c \times fl}$. Let the channels $\bm{x}_{0}, \cdots, \bm{x}_{c'}$ denote the target variables and $\bm{x}_{c'+1}, \cdots, \bm{x}_{c}$ denote all exogenous variables with their future values known. 
First, we replace the forecast values for the exogenous channels with the \textit{true} future values ($\bm{Y}$) and transpose it:
$
    \hat{\bm{Y}}_e = 
    \left[
            \hat{\bm{y}}_0,
            \cdots,
            \hat{\bm{y}}_{c'},
            \bm{y}_{c'+1},
            \cdots,
            \bm{y}_{c}
    \right] \in \mathbb{R}^{fl \times c}
$.
Next, to learn inter-channel \textit{lagged} correlations, we patch $\hat{\bm{Y}}_e$ into a series of overlapped windows (\ie patching with stride$=1$) to create a new tensor: $\hat{\bm{Y}}_{e,p} \in \mathbb{R}^{fl \times \Delta \times c}$, where $\Delta = 2 \cdot l + 1$ with $l$ being the context length to incorporate on either side of a time point\footnote{This needs padding $\hat{\bm{Y}}_e$ with zeros of length $l$ on both sides.}.
Subsequently, we pass $\hat{\bm{Y}}_{e,p}$ through a vanilla TSMixer block with channel mixing enabled.
Thus, the lagged dependency of the forecasts for the target channels on the exogenous channels is seamlessly learned.
Finally, we attach a linear head to produce the forecasts for the target channels which is then reshaped as $\hat{\bm{Y}} \in \mathbb{R}^{c' \times fl}$. Thus, TTM easily handles exogenous infusion which is a practical requirement in any industrial forecasting problem.
Figure~\ref{fig:ttm_overview}(c) depicts this procedure.

\section{Experiments and Results}

\subsection{Datasets \& Metrics }
\label{sec:data_info}

Pre-training employs a subset of $\sim$1B samples from  Monash~\cite{monash} and Libcity~\cite{libcity} data collection. 
We specifically exclude a few datasets (like yearly, monthly) as they do not possess sufficient length for the long-term forecasting task.
Moreover, we remove all the datasets that we utilize for evaluation (\ie Weather, Electricity, and Traffic).
For zero/few-shot evaluation we consider seven public datasets (\textbf{D1}): ETTH1, ETTH2, ETTM1, ETTM2, Weather, Electricity, and Traffic as popularly used in most prior state-of-the-art (SOTA) works~\cite{informer,patchtst}.
Since these datasets do \textit{not} contain any exogenous variables nor exhibit cross-channel correlation benefits, we incorporate four other datasets (\textbf{D2}) for separately validating the efficacy of the decoder channel mixing and exogenous mixer module: bike sharing~(BS)~\cite{misc_bike_sharing_dataset_275}, carbon capture plant~(CC)~\cite{carbon}, and 2 more datasets, Application (APP) and Service (SER), from Business and IT observability domain~\cite{ITBiz_data,automixer}. For full details, refer to the Appendix~\ref{appendix:data_info}. We use mean squared error (MSE) as the standard error metric. In addition, we use the following relative improvement metrics: (i) forecast improvement percentage (\textit{f-imp(\%)}) which refers to the MSE (\%) improvement of TTM over the considered baseline, averaged across all datasets, and (ii) size improvement metric (\textit{s-imp(X)})
is calculated as the ratio of the baseline model size to the TTM model size (i.e., total parameters).

\subsection{SOTA Benchmarks} We benchmark\footnote{For all tables, we highlight the best and second best models with \textbf{bold} and \uline{underline}, respectively. We denote TTM's improvement and degradation \wrt a baseline with $\uparrow$ and $\downarrow$ respectively.} TTM with 24 of the latest open-sourced SOTA forecasting models categorized as follows:
\begin{enumerate*}[label=(\alph*)]
    \item \textbf{TS pre-trained models:} Lag-Llama~\cite{lagllama}, TimesFM~\cite{timesfm}, Moirai~\cite{moirai}, Chronos~\cite{chronos} and Moment~\cite{moment}.
    \item \textbf{LLM-based TS pre-trained models:} GPT4TS~\cite{gpt4ts}, LLMTime~\cite{llmtime}, Time-LLM~\cite{timellm}, UniTime~\cite{unitime}
     \item \textbf{Self-supervised pre-trained models}: SimMTM~\cite{dong2023simmtm},Ti-MAE~\cite{timae}, TST~\cite{tst}, LaST~\cite{laST}, TF-C~\cite{tfc}, CoST~\cite{CoST} and Ts2Vec~\cite{ts2vec}
    \item \textbf{Other architectures:} PatchTST~\cite{patchtst}, TSMixer~\cite{tsmixer}, TimeMixer~\cite{timemixer}, iTransformer~\cite{itransformer}, DLinear~\cite{dlinear} and TimesNet~\cite{timesnet}, FEDFormer~\cite{fedformer} and Autoformer~\cite{autoformer}. 
\end{enumerate*}

\subsection{TTM Model Details}
\label{appendix:ttm_implementation details}
We pre-train three primary variants of TTM as follows: (i) \textbf{TTM-Base (\ttms)}: 1M parameter model trained with context length, $sl=512$ and patch length, $pl=64$, (ii) \textbf{TTM-Enhanced (\ttmm)}: 4M parameter model trained with $sl=1024$ and $pl=128$, (iii) \textbf{TTM-Advanced (\ttml)}: 5M parameter model trained with $sl=1536$ and $pl=128$. These TTMs are pre-trained using the 1B pre-training dataset, which takes only 24-30 hours with 6 A100 GPUs, a notably faster time compared to existing counterparts which often take days to weeks. Additionally, for secondary studies, we utilize \textbf{Quick TTM (\ttmq)}, a variant trained on a smaller subset of the Monash dataset ($\sim$250 million samples), requiring only 4-6 hours for pre-training. 

Although, a TTM model needs to be pre-trained for a specific forecast length (FL), we provide two forecast length adaption (FLA) techniques (explained in Section~\ref{sec:fla}) that enable a pre-trained TTM to work across different FLs. Users can either build a direct pre-trained model (from one of the above variants) targeting a specific FL, or use the FLA techniques to adapt an existing TTM model to their application setting.
Primary results are reported using the direct approach, and a detailed ablation study is provided to compare the effectiveness of various FLA techniques. In the direct approach, model parameter size varies across FLs and we report the average parameter size in the result tables. 
TTM fine-tuning and inferencing are highly efficient and fast, requiring only 1 GPU or even CPU execution. 
All model hyperparameters are chosen based on validation performance, and final test results are reported.
Refer to Appendix~\ref{app:sec:TTM Model Hyper-parameters and Baselines} for detailed model specifications and hyper-parameters.

\begin{figure*}[h]
  \begin{minipage}{1\textwidth}

    \centering
    \setlength{\tabcolsep}{1.1pt}
    \scalebox{.8}{

  \begin{tabular}{|c|c|c|c|c|c|c|c|} \hline 
   \centering
 \makecell{Data} & \makecell{\ttms~} & \makecell{\ttmm~} & \makecell{\ttml~} & \makecell{\moirais~} & \makecell{\moiraib~} & \makecell{\moirail~} & \makecell{TimesFM}\\ \hline 
ETTH1 & \textbf{0.394} & 0.404 & \uline{0.4} & \uline{0.4} & 0.434 & 0.51 & 0.479 \\ 
ETTH2 & 0.345 & \uline{0.335} & \textbf{0.333} & 0.341 & 0.346 & 0.354 & 0.403 \\ 
ETTM1 & 0.386 & \uline{0.38} & \textbf{0.362} & 0.448 & 0.382 & 0.39 & 0.429 \\ 
ETTM2 & 0.281 & \uline{0.271} & \textbf{0.252} & 0.3 & 0.272 & 0.276 & 0.334 \\ 
Weather & \uline{0.237} & 0.238 & \textbf{0.231} & 0.242 & 0.238 & 0.26 & - \\ 
Electricity & 0.205 & 0.194 & \uline{0.192} & 0.233 & \textbf{0.188} & \textbf{0.188} & - \\ 
\hline 
\multicolumn{1}{|c|}{\makecell{\textbf{Size}}} & \textbf{1M} & \textbf{4M} & \textbf{5M} & \textbf{14M} & \textbf{91M} & \textbf{311M} & \textbf{200M}\\ \hline 
\multicolumn{4}{|c|}{\makecell{\textbf{\ttms~} \textit{f-imp(\%) s-imp(X)}}} & \textbf{6\% $\uparrow$ 14X $\uparrow$} & \textbf{1\% $\downarrow$ 91X $\uparrow$} & \textbf{4\% $\uparrow$ 311X $\uparrow$} & \textbf{15\% $\uparrow$ 200X $\uparrow$} \\ 
\multicolumn{4}{|c|}{\makecell{\textbf{\ttmm~} \textit{f-imp(\%) s-imp(X)}}} & \textbf{7\% $\uparrow$ 4X $\uparrow$} & \textbf{1\% $\uparrow$ 23X $\uparrow$} & \textbf{6\% $\uparrow$ 78X $\uparrow$} & \textbf{16\% $\uparrow$ 50X $\uparrow$} \\ 
\multicolumn{4}{|c|}{\makecell{\textbf{\ttml~} \textit{f-imp(\%) s-imp(X)}}} & \textbf{10\% $\uparrow$ 3X $\uparrow$} & \textbf{4\% $\uparrow$ 18X $\uparrow$} & \textbf{9\% $\uparrow$ 62X $\uparrow$} & \textbf{19\% $\uparrow$ 40X $\uparrow$} \\ 
\hline 
 \end{tabular}
     }
\captionof{table}{\textbf{Zero-shot} forecast-improvement \textit{(f-imp)} and model size-improvement \textit{(s-imp)} of TTM over Moirai (ICML'24) and TimesFM (ICML'24).
MSE averaged across  \textsc{\emph{FL}}~$\in \{96,192,336,720\}$. Electricity and Weather results for TimesFM are not reported as its used by TimesFM for pretraining. Similarly, Traffic was used in pre-training for both Moirai and TimesFM. Full table in Appendix ~\ref{appendix:zs}}
\label{tab:n_zs_moirai_avg}

  \end{minipage}%
   
\end{figure*}

\begin{figure*}[h]
  \begin{minipage}{1\textwidth} 
    \centering
    \setlength{\tabcolsep}{1.1pt}
    \scalebox{.8}{
    \begin{tabular}{|c|c|c|c|c|c|c|c|c|} \hline 
     \centering
 \makecell{Data} & \makecell{\ttms~} & \makecell{\ttmm~} & \makecell{\ttml~} & \makecell{\chronust~} & \makecell{\chronuss~} & \makecell{\chronusb~} & \makecell{\chronusl~} & \makecell{Lag-llama}\\ \hline 
ETTH1 & \textbf{0.204} & 0.227 & \uline{0.214} & 0.311 & 0.302 & 0.252 & 0.266 & 0.334 \\ 
ETTH2 & \textbf{0.131} & \uline{0.151} & 0.162 & 0.177 & 0.16 & 0.164 & 0.155 & 0.168 \\ 
ETTM1 & \uline{0.206} & 0.239 & \textbf{0.19} & 0.839 & 0.486 & 0.49 & 0.538 & 0.842 \\ 
ETTM2 & \uline{0.124} & 0.128 & \textbf{0.117} & 0.206 & 0.174 & 0.19 & 0.187 & 0.308 \\ 
Weather & 0.039 & \uline{0.032} & 0.043 & 0.043 & 0.046 & \textbf{0.03} & 0.033 & 0.126 \\ 
Electricity & \textbf{0.335} & 0.351 & 0.349 & 0.423 & 0.377 & 0.344 & \uline{0.339} & 0.393 \\ 
Traffic & 0.246 & \textbf{0.24} & 0.244 & 0.291 & 0.3 & 0.28 & 0.269 & \uline{0.243} \\ 
\hline 

\multicolumn{1}{|c|}{\makecell{\textbf{Size}}} & \textbf{1M} & \textbf{4M} & \textbf{5M} & \textbf{8M} & \textbf{46M} & \textbf{201M} & \textbf{709M} & \textbf{3M}\\ \hline 
\multicolumn{4}{|c|}{\makecell{\textbf{\ttms~} \textit{f-imp(\%) s-imp(X)}}} & \textbf{32\% $\uparrow$ 8X $\uparrow$} & \textbf{26\% $\uparrow$ 46X $\uparrow$} & \textbf{17\% $\uparrow$ 201X $\uparrow$} & \textbf{18\% $\uparrow$ 709X $\uparrow$} & \textbf{40\% $\uparrow$ 3X $\uparrow$} \\ 
\multicolumn{4}{|c|}{\makecell{\textbf{\ttmm~} \textit{f-imp(\%) s-imp(X)}}} & \textbf{30\% $\uparrow$ 2X $\uparrow$} & \textbf{24\% $\uparrow$ 12X $\uparrow$} & \textbf{15\% $\uparrow$ 50X $\uparrow$} & \textbf{16\% $\uparrow$ 177X $\uparrow$} & \textbf{37\% $\uparrow$ 1X $\downarrow$} \\ 
\multicolumn{4}{|c|}{\makecell{\textbf{\ttml~} \textit{f-imp(\%) s-imp(X)}}} & \textbf{28\% $\uparrow$ 2X $\uparrow$} & \textbf{22\% $\uparrow$ 9X $\uparrow$} & \textbf{12\% $\uparrow$ 40X $\uparrow$} & \textbf{13\% $\uparrow$ 142X $\uparrow$} & \textbf{37\% $\uparrow$ 2X $\downarrow$} \\ 
\hline 
 \end{tabular}
   }
    \captionof{table}{{\textbf{Zero-shot} forecast-improvement \textit{(f-imp)} and model size-improvement \textit{(s-imp)} of TTM over Chronos and Lag-llama over the last test-window. Since Chronos and Lag-llama recommend/report results with shorter forecast lengths, we use  \textsc{\emph{FL}}~$\in \{24,48,60,96,192\}$. Mean MSE across FLs is reported. Full table in the Appendix~\ref{appendix:zs}}}
    
    \label{tab:n_zs_2_avg}

  \end{minipage}%

\end{figure*}

\begin{figure*}[h]
  \begin{minipage}{1\textwidth}

    \centering
    \setlength{\tabcolsep}{2pt}
    \scalebox{.75}{
        \begin{tabular}{|c|c|c|c|c|}
            \hline
            Model        & \makecell{GPU TIME \textbf{(ms)}}   & \makecell{Params  \textbf{(M)}}      & \makecell{MEM \textbf{(GB)}}  & \makecell{CPU TIME \textbf{(s)}} \\ \hline
            \makecell{\textbf{\ttms}}                      & \textbf{4.7 }             & \textbf{0.8  }          & \textbf{0.06   }        &  \textbf{0.01}  \\   \hline
            \makecell{\chronusb \\ \small{(2024)} }      & \makecell{1395\\	(298X)}  & \makecell{201 \\(251X)} & \makecell{16 \\(267X)}  & \makecell{2340\\	(239KX)} \\  \hline
            \makecell{\chronusl \\ \small{(2024)}}       & \makecell{1393\\	(298X)}  & \makecell{709\\ (886X)} & \makecell{41 \\(683X)}  & \makecell{2352\\ (240KX)} \\  \hline
            \makecell{\chronuss \\ \small{(2024)} }      & \makecell{1386\\	(296X)}  & \makecell{46 \\(58X)}   & \makecell{6 \\(100X)}   & \makecell{2349\\	(240KX)}\\  \hline
            \makecell{\chronust \\ \small{(2024)} }      & \makecell{1389\\	(297X)}  & \makecell{8 \\(10X)}    & \makecell{2 \\(33X)}    & \makecell{2504\\	(256KX)}\\  \hline
            \makecell{GPT4TS \\ \small{(NeurIPS '23)} }  & \makecell{\underline{13.9}\\	(3X)}  & \makecell{87\\(109X)} & \makecell{1.34\\(36X)} & \makecell{\underline{0.3}\\ (26X)}\\  \hline
            \makecell{Lag-Llama \\ \small{(2024)}}       & \makecell{1619\\	(346X)}  & \makecell{\underline{2.4} \\(3X)}   & \makecell{0.2 \\(3X)}   & \makecell{37.5\\(3830X)} \\  \hline
            \makecell{\moirais \\ \small{(ICML '24)} }   & \makecell{205\\	(44X)}   & \makecell{14 \\(18X)}   & \makecell{\underline{0.1} \\(2X)}   & \makecell{1.4\\(141X)}\\  \hline
            \makecell{\moirail \\ \small{(ICML '24)}}    & \makecell{693\\	(148X)}  & \makecell{311 \\(389X)} & \makecell{2 \\(33X)}    & \makecell{10.5\\(1070X)}\\  \hline
            \makecell{\moiraib \\ \small{(ICML '24)}  }  & \makecell{335\\	(72X)}   & \makecell{91 \\(114X)}  & \makecell{1 \\(17X)}    & \makecell{4.1\\(421X)}\\  \hline
            \makecell{Moment-L \\ \small{(ICML '24)}}    & \makecell{88\\	(19X)}   & \makecell{348 \\(435X)} & \makecell{8 \\(133X)}   & \makecell{1.4\\(144X)}\\ \hline
            \makecell{TimesFM \\ \small{(ICML '24)}}     & \makecell{24\\	(5X)}    & \makecell{200 \\(250X)} & \makecell{2 \\ (33X)}   & \makecell{0.4\\(46X)}\\ \hline
        \end{tabular}        

     }
    \captionof{table}{{\textbf{Computational improvement} of TTM \wrt existing TS pre-trained models. Inference time per-batch in GPU and CPU, total parameters (Params), and maximum GPU memory usage (MEM) are reported. nX indicates the scaling factor for TTM's improvement. Set-up details are in the Appendix~\ref{appendix:ttm_comp_2}}}
    \label{tab:n_timing}
  \end{minipage}%
\vspace{-0.5cm}
\end{figure*}

\subsection{TTM's Zero-shot Performance and Inference Cost}
Recently, popular pre-trained models like TimesFM, Moirai, Chornos, Lag-llama, and LLMTime have gained traction for their zero-shot (ZS) forecasting capabilities. Among these, Chornos, Lag-llama, and LLMTime suffer from lengthy ZS inference time, posing practical challenges for testing across all sliding windows of the test set. To address this, LLMTime suggests using the last test window for benchmarking, a practice we also adopt for comparing with this set of SOTA models. On the other hand, TimesFM and Moirai exhibit comparatively faster ZS inference speeds, enabling testing across all sliding windows of the test set. Table~\ref{tab:n_zs_moirai_avg} presents a comparison of TTM performance with Moirai and TimesFM. Despite having significantly fewer parameters, the variants of TTM consistently outperform most benchmark variants. Notably, \ttml, which is 3-62X smaller than all Moirai variants and 40X smaller than TimesFM, outperforms the Moirai variants by 4-10\% and TimesFM by 19\%. 
Even \ttms, with just 1M parameters, outperforms most benchmarks by a considerable margin, highlighting the effectiveness of TTM. Moreover, as depicted in Appendix \ref{appendix:zs_fullshot_short}, TTM zero-shot results consistently outperform the full-shot results of popular architectures in short context length settings. 
Likewise, Table~\ref{tab:n_zs_2_avg} presents a comparison of TTM performance with Chronos and Lag-llama on the last test-window set. As indicated, \ttms~which is 8-709X smaller than Chronos, outperforms it by 17-32\%. Likewise \ttms, which is 2-3X smaller than Lag-llama, outperforms it by 40\%. In addition, TTM also outperforms the massive LLMTime and UniTime by over 25\% as reported in Appendix~\ref{appendix:llmtime_unitime}.
Table \ref{tab:n_timing} presents the inference time per batch and maximum GPU memory requirement of different TS pre-trained models. Notably, TTM exhibits the lowest inference time and memory usage among them. 



\subsection{TTM's Few-shot and Full-shot Head Probing Performance}

In operational deployments, users typically leverage a small set of target data for fine-tuning to enhance the model performance. 
In this regard, TTM provides a highly efficient quick fine-tuning process, enabling users to enhance forecasting accuracy swiftly by training only the model head. 
GPT4TS and Time-LLM are two state-of-the-art pre-trained models that present results for few-shot training. 
As demonstrated in Table~\ref{tab:n_fs5_avg}, \ttms~surpasses GPT4TS by 15\% and Time-LLM by 10\% in the few-shot 5\% setting, where only 5\% of the train data is used for fine-tuning.
In addition, we also report the Few-shot 5\% results of several popular SOTA architectures in Table~\ref{tab:n_fs5_avg}, where TTM demonstrates superior performance. This underscores the significance of TTM's pre-trained weights, which substantially contribute to its effectiveness in data-constrained scenarios. 
Likewise, TTM also excels in few-shot cross-transfer learning tasks outperforming popular SOTAs (including SimMTM~\cite{dong2023simmtm}) as shown in the Appendix~\ref{appendix:cross_transfer}.


Alternatively, if the train split of the complete target dataset is available, head probing using the entire dataset becomes feasible.
This involves fine-tuning the model head using all available data while keeping the backbone weights unchanged. Recently, the Moment~\cite{moment} model has achieved the SOTA results in head probing as compared to GPT4TS and Time-LLM. However, as indicated in Table~\ref{tab:n_hp_avg}, TTM further outperforms the results reported by Moment by 3-4\%. In addition, TTM head probing results are very competitive as compared to the full end-to-end training of popular architectures as depicted in Appendix~\ref{appendix:hp_full_shot_512}. Hence, TTM, with its significantly reduced model size and the absence of compute-intensive components like self-attention, enables quick fine-tuning of models compared to the cumbersome process required by the massive Transformer models. 
Note that Moment is excluded from the comparison of zero/few-shot forecasting results because it does not report them.


\begin{figure*}[h]
  \begin{minipage}{0.95\textwidth}
    \centering
    \setlength{\tabcolsep}{1.5pt}
    \scalebox{.68}{

   \begin{tabular}{|c|c|c|c|c|c|c|c|c|c|c|c|} \hline 
   \makecell{} & \multicolumn{5}{|c|}{\makecell{\textbf{Pre-trained models fine-tuned on 5\% data}}} & \multicolumn{6}{|c|}{\makecell{\textbf{Other popular architectures trained on 5\% data}}} \\ \hline 

   \makecell{Data \\ } & \makecell{\textbf{\ttms~} \\ (Ours)} & \makecell{\textbf{\ttmm~} \\ (Ours)} & \makecell{\textbf{\ttml~} \\ (Ours)} & \makecell{\textbf{GPT4TS} \\ (NeurIPS '23)} & \makecell{\textbf{Time-LLM} \\ (ICLR '24)} & \makecell{PatchTST \\ (ICLR '23)} & \makecell{TSMixer \\ (KDD '23)} & \makecell{TimeMixer \\ (ICLR '24)} & \makecell{iTransformer \\ (ICLR '24)} & \makecell{TimesNet \\ (ICLR '23)} & \makecell{Dlinear \\ (AAAI '23)}\\ \hline


ETTH1 & \textbf{0.383} & \uline{0.385} & 0.386 & 0.682 & 0.627 & 0.695 & 0.635 & 1.088 & 0.756 & 0.926 & 0.75 \\ 
ETTH2 & 0.324 & \uline{0.318} & \textbf{0.314} & 0.401 & 0.382 & 0.439 & 0.385 & 0.508 & 0.437 & 0.464 & 0.828 \\ 
ETTM1 & \uline{0.376} & 0.378 & \textbf{0.361} & 0.472 & 0.425 & 0.526 & 0.479 & 0.578 & 0.568 & 0.717 & 0.401 \\ 
ETTM2 & 0.272 & \uline{0.268} & \textbf{0.253} & 0.308 & 0.274 & 0.314 & 0.297 & 0.34 & 0.309 & 0.344 & 0.399 \\ 
Weather & \uline{0.234} & 0.24 & \textbf{0.229} & 0.263 & 0.261 & 0.27 & 0.268 & 0.317 & 0.297 & 0.298 & 0.264 \\ 
Electricity & 0.183 & 0.207 & 0.18 & 0.178 & \uline{0.177} & \textbf{0.176} & 0.197 & 0.239 & 0.202 & 0.402 & \uline{0.177} \\ 
Traffic & 0.433 & 0.437 & 0.49 & 0.434 & \uline{0.423} & \textbf{0.418} & 0.435 & 0.503 & 0.452 & 0.867 & 0.451 \\ 
\hline 
\multicolumn{1}{|c|}{\makecell{\textbf{Size}}} & \textbf{1M} & \textbf{4M} & \textbf{5M} & \textbf{84M} & \textbf{7B} & \textbf{} & \textbf{} & \textbf{} & \textbf{} & \textbf{} & \textbf{}\\ \hline 
\multicolumn{4}{|c|}{\makecell{\textbf{\ttms~} \textit{f-imp(\%) s-imp(X)}}} & \textbf{15\% $\uparrow$ 84X $\uparrow$} & \textbf{10\% $\uparrow$ 7KX $\uparrow$} & \textbf{17\% $\uparrow$ } & \textbf{15\% $\uparrow$ } & \textbf{31\% $\uparrow$ } & \textbf{22\% $\uparrow$ } & \textbf{40\% $\uparrow$ } & \textbf{23\% $\uparrow$ } \\ 
\multicolumn{4}{|c|}{\makecell{\textbf{\ttmm~} \textit{f-imp(\%) s-imp(X)}}} & \textbf{13\% $\uparrow$ 21X $\uparrow$} & \textbf{8\% $\uparrow$ 1.7KX $\uparrow$} & \textbf{15\% $\uparrow$ } & \textbf{13\% $\uparrow$ } & \textbf{30\% $\uparrow$ } & \textbf{20\% $\uparrow$ } & \textbf{40\% $\uparrow$ } & \textbf{21\% $\uparrow$ } \\ 
\multicolumn{4}{|c|}{\makecell{\textbf{\ttml~} \textit{f-imp(\%) s-imp(X)}}} & \textbf{15\% $\uparrow$ 17X $\uparrow$} & \textbf{11\% $\uparrow$ 1.4KX $\uparrow$} & \textbf{17\% $\uparrow$ } & \textbf{15\% $\uparrow$ } & \textbf{32\% $\uparrow$ } & \textbf{22\% $\uparrow$ } & \textbf{41\% $\uparrow$ } & \textbf{23\% $\uparrow$ } \\ 
\hline 
 \end{tabular}
 
   }
   \vspace{-.2cm}
    \captionof{table}{{\textbf{Few-shot 5\%}. MSE averaged across  \textsc{\emph{FL}}~$\in \{96, 192, 336, 720\}$, models are trained with 5\% train data (Appendix~\ref{appendix:fs5}).}}
    \label{tab:n_fs5_avg}
  \end{minipage}%
\end{figure*}


\begin{figure*}[h]
  \begin{minipage}{1\textwidth}
    \centering
    \setlength{\tabcolsep}{1.5pt}
    \scalebox{.8}{

\begin{tabular}{|c|c|c|c|c|c|c|} \hline 
 \makecell{Data \\ } & \makecell{\ttms~ \\ (Ours)} & \makecell{\ttmm~ \\ (Ours)} & \makecell{\ttml~ \\ (Ours)} & \makecell{Moment \\ (ICML '24)} & \makecell{GPT4TS \\ (NeurIPS '23)} & \makecell{Time-LLM \\ (ICLR '24)}\\ \hline 
ETTH1 & \textbf{0.398} & 0.406 & \uline{0.402} & 0.42 & 0.426 & 0.466 \\ 
ETTH2 & \uline{0.33} & 0.338 & \textbf{0.327} & 0.346 & 0.346 & 0.342 \\ 
ETTM1 & 0.355 & 0.35 & \textbf{0.338} & \uline{0.349} & 0.354 & 0.41 \\ 
ETTM2 & \uline{0.257} & \textbf{0.252} & 0.264 & 0.266 & 0.275 & 0.273 \\ 
Weather & \uline{0.234} & 0.239 & \textbf{0.233} & \uline{0.234} & 0.244 & - \\ 
Electricity & 0.164 & \uline{0.161} & \textbf{0.16} & 0.174 & 0.172 & - \\ 
Traffic & 0.399 & \uline{0.398} & \textbf{0.385} & 0.42 & 0.419 & - \\ 
\hline 
\multicolumn{1}{|c|}{\makecell{\textbf{Size}}} & \textbf{1M} & \textbf{4M} & \textbf{5M} & \textbf{348M} & \textbf{84M} & \textbf{7B}\\ \hline 
\multicolumn{4}{|c|}{\makecell{\textbf{\ttms~} \textit{f-imp(\%) s-imp(X)}}} & \textbf{3\% $\uparrow$ 348X $\uparrow$} & \textbf{4\% $\uparrow$ 84X $\uparrow$} & \textbf{9\% $\uparrow$ 7000X $\uparrow$} \\ 
\multicolumn{4}{|c|}{\makecell{\textbf{\ttmm~} \textit{f-imp(\%) s-imp(X)}}} & \textbf{3\% $\uparrow$ 87X $\uparrow$} & \textbf{4\% $\uparrow$ 21X $\uparrow$} & \textbf{9\% $\uparrow$ 1750X $\uparrow$} \\ 
\multicolumn{4}{|c|}{\makecell{\textbf{\ttml~} \textit{f-imp(\%) s-imp(X)}}} & \textbf{4\% $\uparrow$ 70X $\uparrow$} & \textbf{6\% $\uparrow$ 17X $\uparrow$} & \textbf{10\% $\uparrow$ 1400X $\uparrow$} \\ 

\hline 
 \end{tabular}
   
   }
    \captionof{table}{{\textbf{Full-shot head probing:} Finetuning the pre-trained model heads on full data with backbone weights frozen. MSE averaged across  \textsc{\emph{FL}} 96, 720 as reported in \cite{moment}. Time-LLM results for large datasets are not reported in \cite{moment} due to computational challenges (Appendix\ref{appendix:hp_full_shot_512}).}}
    \label{tab:n_hp_avg}
\end{minipage}%
  
\end{figure*}

\begin{figure*}[h]

\begin{minipage}{1\textwidth}

      \centering%
    \setlength{\tabcolsep}{2.5pt}
    \scalebox{0.8}{

    \begin{tabular}{|c|c|c|c|c|c|c|}
    \hline
     Models & \makecell{BS} & \makecell{CC} & \makecell{APP} & \makecell{SER} & \makecell{\textit{f-imp\%}}\\ \hline 
    
    \ttmq & \uline{0.635} & \uline{0.261} & 0.073 & 0.143 & 18\%\\ 
    PatchTST & 0.735 & 0.267 & 0.060 & 0.119 & 15\%\\ 
    TSMixer-CC & 0.651 & 0.284 & \uline{0.053} & 0.136 & 15\% \\ 
    TSMixer-CM & 0.716 & 0.303 & 0.069 & \uline{0.118} & 20\% \\ 
    TSMixer & 0.664 & 0.267 & 0.066 & 0.134 & 17\% \\ 
    GPT4TS & 0.645 & \uline{0.254} & 0.075 & 0.135 & 18\%\\ 
    \textbf{\ttmq-CM} & \textbf{0.582} & \textbf{0.250} & \textbf{0.042} & \textbf{0.114} & \\ \hline
    
    \hline 
    
    \hline 
    \end{tabular}}

    \captionof{table}{{\textbf{Effect of decoder mixing and exog.\ fusion.} MSE results are reported using $(sl, fl)$ with values of $(512, 96)$ for BS dataset and $(96, 24)$ for other D2 datasets.}} 
    \label{tab:exog_mixer_avg}
    
  \end{minipage}%
  
\end{figure*}


\begin{figure*}[h]

  \begin{minipage}{.48\textwidth}
  \centering
     \includegraphics[width=0.7\linewidth]{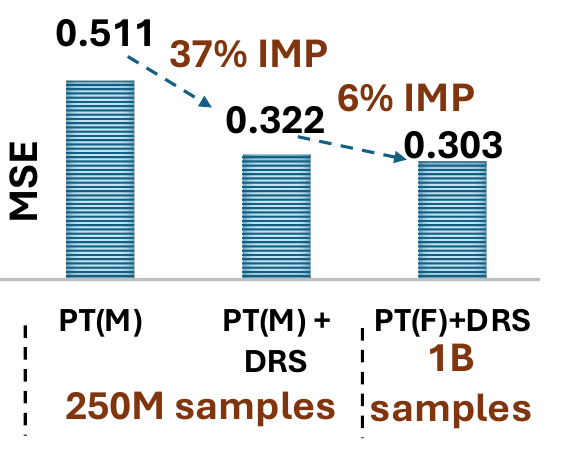}
    \caption{{\textbf{Impact of pre-training data (PT) and diverse resolution sampling (DRS) technique.} PT(M): pretraining with only Monash data. PT(F): full pretraining data used. Average MSE of zero-shot results across  \textsc{\emph{FL}} 96, 192 reported.}}
    \label{fig:n_data_abl}
  \end{minipage}%
    \hspace{0.2cm}
  \begin{minipage}{0.48\textwidth}
  \centering
    \vspace{0.5cm}
    \includegraphics[width=0.7\linewidth]{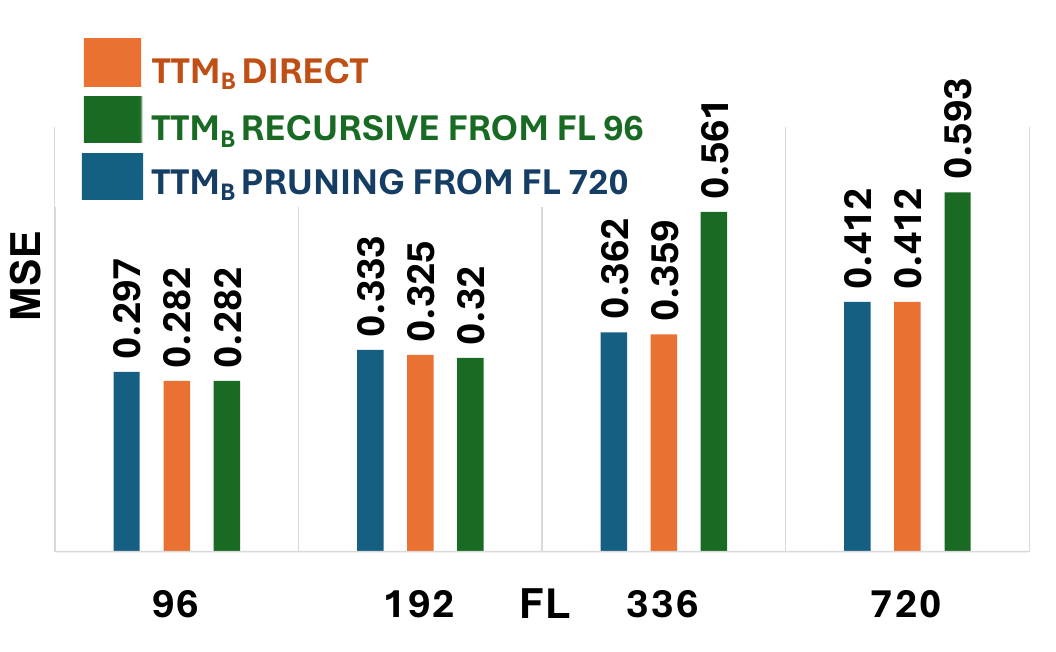}
    \caption{{\textbf{FL adaptation:} impact of adapting \ttms~( \textsc{\emph{FL}} $720$) and \ttms~ (\textsc{\emph{FL}} $96$) to all other FLs. MSE averaged across all D1 datasets is reported for  \textsc{\emph{FL}}~$\in \{96,192,336,720\}$. Best viewed in color.}}
    \label{fig:n_adap_abl}
  \end{minipage}
\end{figure*}

\begin{figure*}[h]

  \begin{minipage}{0.48\textwidth}
    \centering
    \vspace{0.5cm}
    \setlength{\tabcolsep}{1pt}
    \scalebox{.8}{
   \begin{tabular}{|c|c|c|c|c|} \hline 
 \multicolumn{1}{|l|}{} & \multicolumn{2}{c|}{\makecell{Less   PT Data \\ (250M)}} & \multicolumn{2}{c|}{\makecell{More   PT Data \\ (1B)}} \\ \hline 

Data & w/o AP               & w/ AP              & w/o RPT             & w/ RPT            \\ \hline 
ETTH1                  & 0.369                & \textbf{0.365}              & 0.366               & \textbf{0.364}             \\
ETTH2                  & \textbf{0.283}                & 0.285              & 0.285               & \textbf{0.277}             \\
ETTM1                  & 0.446                & \textbf{0.413}              & 0.341               & \textbf{0.322}             \\
ETTM2                  & 0.191                & \textbf{0.187}              & 0.18                & \textbf{0.171}             \\
Weather                & 0.159                & \textbf{0.154}              & 0.\textbf{153 }              & 0.158             \\
Electricity            & 0.179                & \textbf{0.169}              & 0.178               & \textbf{0.166 }            \\
Traffic                & 0.521                & \textbf{0.518}              & 0.528               & \textbf{0.514 }            \\ \hline 
\textbf{IMP (\%)}               & \multicolumn{2}{c|}{\textbf{3\%}}                   & \multicolumn{2}{c|}{\textbf{3\%}}                \\ \hline 
 \end{tabular}
   }
    \captionof{table}{{\textbf{Impact of AP and RPT:} Impacts of adaptive patching (AP) in less pre-training (PT) data setting (i.e., \ttmq), and resolution prefix tuning (RPT) in more pre-training (PT) data setting (i.e., \ttms). Zero-shot results on FL 96 reported. [`w/': with, `w/o': `without'.]}}
    \label{tab:n_abl_ap_freq}
  \end{minipage}%
    \hspace{0.3cm}
   \begin{minipage}{0.48\textwidth}

    \centering
\includegraphics[width=0.8\linewidth]{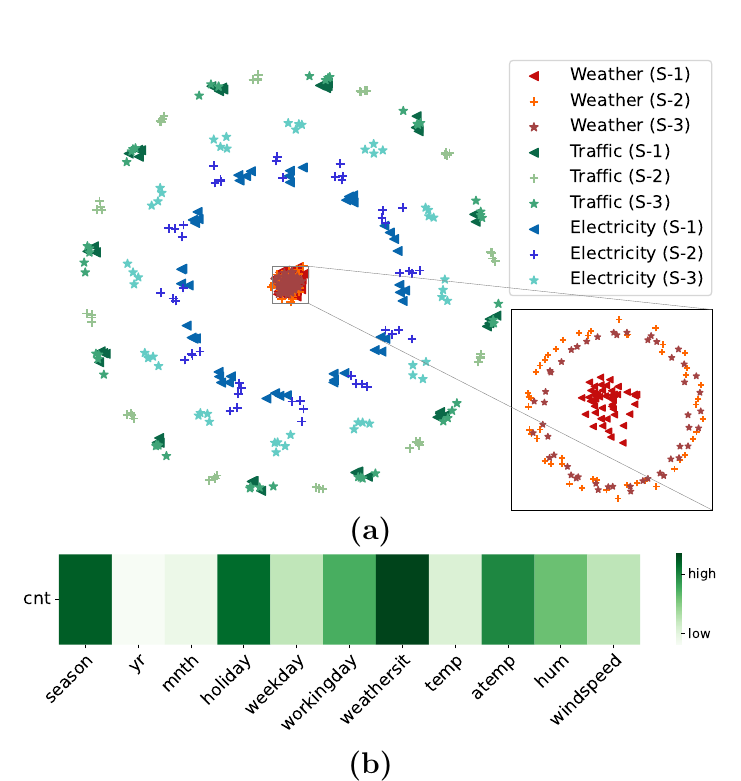}
\caption{(a) \textbf{TTM embedding projections} across 3 datasets and 3 segments within datasets.\label{fig:ttm-embedding} 
(b) Cross-channel \textbf{attention based explanation}.\label{fig:bikesharing-exog-explain}}

  \end{minipage}%

\end{figure*}

\subsection{TTM's Effectiveness in Cross-channel and Exogenous Modeling}
Since the datasets (\textbf{D1}) used in previous experiments do not have exogenous variables, we evaluate the effectiveness of TTM on  4 other datasets (\textbf{D2}, as explained in Section~\ref{sec:data_info}) to quantify its benefits. Since these datasets are already very small, we used their full data for fine-tuning. Table~\ref{tab:exog_mixer_avg} shows the performance of the pre-trained \ttmq~model fine-tuned on the target data with exogenous mixer module and decoder channel-mixing enabled (TTM-CM). We compare TTM-CM with plain TTM finetuning and other  primary SOTAs (PatchTST, TSMixer variants, and GPT4TS) trained from scratch. Specifically, we compare with TSMixer with channel-mixing enabled (TSMixer-CM) and TSMixer with cross-channel reconciliation head (TSMixer-CC)~\cite{tsmixer} as they are the latest SOTAs in channel-correlation modelling. From Table~\ref{tab:exog_mixer_avg}, we can see that TTM-CM outperforms all the competitive models with a significant margin (15-44\%), thus, demonstrating the power of TTM in capturing inter-channel correlations.



\subsection{Ablation Studies}\label{sec:Ablation Studies} The impacts of various techniques used in TTM are analyzed here. \\
\textbf{Pre-training data (Quality Vs Quantity):} Figure~\ref{fig:n_data_abl} demonstrates the vital role of both pretraining data and diverse resolution sampling (DRS). 
Initially, the zero-shot results were unsatisfactory when pre-training TTM with the smaller Monash dataset (i.e., PT(M)).
To improve performance, we introduced the DRS technique on the Monash data to
increase diversity and coverage (250M PT samples). 
This significantly improved the results by 37\%. 
In addition, extending the dataset size from 250M to 1B further improved the results by 6\%. 
These experiments highlight that while the quantity of pre-training data is significant, the quality of the data, especially in terms of resolution diversity and coverage, is even more crucial for improving the model performance.


\textbf{Effect of Resolution Prefix Tuning (RPT) and Adaptive Patching(AP): }RPT enhances forecast performance, especially with large and diverse pretraining (PT) data. Adding a learnable resolution prefix token allows models to easily decouple weights across different resolutions, leading to a 3\% improvement in 1B PT data setup (Table~\ref{tab:n_abl_ap_freq}). RPT is also beneficial for very short context length scenarios, improving performance by 8\% (Appendix~\ref{appendix:RPT}). On the other hand, AP generally improves the forecasting performance across all set-ups, but the impact is consistently high in less PT data settings (3\% boost). Further details are in Appendix~\ref{appendix:AP}.

\textbf{Forecast Length Adaptations (FLA):} 
\label{sec:fla}
Given a  \textsc{\emph{FL}},  we can either pre-train a Direct TTM tailored for the specific \textsc{\emph{FL}} or adapt existing TTMs trained on different \textsc{\emph{FL}}s. Two possible adaptations are: (i) \textbf{Pruning:} Take TTM trained on \textsc{\emph{FL}}$'$ where \textsc{\emph{FL}}$'>$\textsc{\emph{FL}},  and prune it to the required \textsc{\emph{FL}} (e.g., TTM $(\textsc{\emph{FL}}=720)$ pruned to other reduced \textsc{\emph{FL}} $\in \{96,192, 336\}$). (ii) \textbf{Recursive:} Take TTM trained on \textsc{\emph{FL}}$'$, where \textsc{\emph{FL}}$'< $\textsc{\emph{FL}} and do recursive prediction (of length \textsc{\emph{FL}}$'$) till we reach the required \textsc{\emph{FL}} (e.g., Extend TTM $(\textsc{\emph{FL}}=96)$ recursively to greater $\textsc{\emph{FL}} \in \{192, 336, 720\}$). Figure~\ref{fig:n_adap_abl} compares these techniques. For shorter adaptation (96 to 192), recursive predictions yield the best performance and match the direct forecast results. However, for wider adaptations (336-96 or 720-96), the pruning approach gives more stable and closer results to the direct forecasts. Hence, using these approaches, TTM models can be easily adapted to various \textsc{\emph{FL}}s based on user requirements.




\subsection{TTM Model Insights \& Explainability}
Figure~\ref{fig:ttm-embedding} illustrates the TTM embeddings from various datasets (weather, traffic, and electricity) using PCA projection, each represented by a different color. From each dataset, three distant, non-overlapping, fixed-length time segments (\textbf{S-1}, \textbf{S-2}, \textbf{S-3}) are selected, each depicted with a unique marker shape. The visualization uses the first and second principal components of the TTM embeddings. The inset image focuses on the weather dataset alone, revealing a deeper structure learned by the TTM architecture. The cyclic orbits in the embeddings reflect the seasonal patterns in the data. Both hourly datasets (traffic and electricity) form concentric orbits due to similar seasonal patterns, while the weather data, with its distinct seasonal pattern, shows cyclic orbits in a different sub-dimension. In addition, the cross-channel attention from the fine-tuned model's channel mixing layers reveals feature importance across channels. As shown in Figure~\ref{fig:bikesharing-exog-explain}, the model focuses on channels like weathersit, season, holiday, and temperature to predict bike-rental counts. These attention model weights correlate with the general data characteristics where bike rental demands are heavily influenced by weather and holidays, providing  explanation for the fine-tuned model predictions.
More details are in Appendix~\ref{appendix:explain}.

\subsection{Discussion on TTM Design choices}
In this section, we intuitively explain the important design choices of TTM that greatly enhance its forecasting accuracy and transfer learning capabilities despite its extremely small model capacity:
\begin{itemize}

    \item All existing pre-trained models use a very high volume of pretraining data (for example, TimesFM used ~300B and Moirai used 27B time-points), hence they naturally require massive model sizes. However, as shown in Figure.\ref{fig:n_data_abl}, we observe that “limited” pretraining data with “high resolution diversity” greatly helps in time-series model generalization, as opposed to simply increasing the pretraining data size. This is an important observation and finding that resolution diversity in pretraining data is very crucial for time-series FMs. Based on these findings, we proceed with a well-reduced dataset (1B samples) with high resolution diversity which naturally reduces our model size compared to counterparts needing to pre-train with several hundred billion time-series. We introduce a high diversity in our data via Diverse Resolution Sampling technique (DRS) which our counterparts fail to do.

    \item Secondly, we opted for TSMixer-based models instead of transformer-based models, which further reduced the model size drastically. The TSMixer architecture has successfully established in the past that interleaving simple gated attentions with mixing components across patches, channels, and features greatly enhances forecasting accuracies with very limited model capacity, as the quadratic time-complexity of self-attentions can be entirely avoided. Following  TSMixer, several other mixer architectures~\cite{timemixer}~\cite{automixer} have been published, reiterating the power of these simple architectures. Thus, avoiding complex transformer architectures further reduced our model size significantly.

    \item In addition, we further increased the modeling power of TSMixer without drastically increasing its size by introducing several innovative components, such as adaptive patching, diverse resolution sampling, and resolution prefix tuning. These enhancements are crucial for effectively handling large pre-training across datasets with varying resolutions, all while keeping the model capacity very minimal.

    \item Finally, framing the pre-training objective as a direct forecasting task demonstrates improved zero-shot performance as compared to the traditional masking-based pre-training approaches. We hypothesize that this method enables the model to effectively learn complex non-linear mappings between the fixed context and forecast windows during pre-training that generalizes well to unseen datasets.
    
\end{itemize}

\section{Conclusions and Future Work}

We propose TTM, an extremely lightweight pre-trained model for multivariate time-series forecasting. Unlike existing large models, TTM is significantly smaller and faster, with efficient pre-training and fine-tuning workflows. Results show that TTM is highly effective in pre-training on heterogeneous datasets despite its limited model capacity. It achieves state-of-the-art results in zero/few-shot forecasting, offering significant computational efficiency while capturing cross-channel relationships and exogenous variables -- critical features lacking in popular methods. Additionally, TTM supports both CPU and GPU deployments, greatly enhancing its adoption and ease of use. Moving forward, we plan to generalize our approach to support other downstream tasks beyond forecasting.

\bibliographystyle{plain}
\bibliography{ttm}
      
\part*{Appendix} 

\appendix

\section{TSMixer Background}
\label{appendix:tsmixer_background}
We employed TSMixer~\cite{tsmixer} as a building block for the proposed TTM model due to its state-of-the-art performance, faster execution, and significantly lower memory usage. However, as explained in the main paper, vanilla TSMixer cannot be trained on multiple diverse datasets. Therefore, it necessitated the incorporation of the proposed novel components.
In this section, we provide a high-level overview of the TSMixer model for a simpler and quicker understanding by the readers.

TSMixer is a lightweight alternative to transformer-based time series models, with no compromise on forecast accuracy. 
TSMixer adopts some well-established pre-processing steps from the literature, such as normalization and patching. 
Additionally, it offers the flexibility of enabling or disabling channel mixing. 
Channel mixing has been found to be beneficial in handling multivariate datasets with cross-channel correlations.
For the main learning process, TSMixer employs a series of MLPMixer~\cite{mlpmixer} blocks that perform inter-patch, intra-patch, and inter-channel mixing operations. 
A mixing operation in TSMixer ensures learning correlations across a specific dimension. 
For example, inter-channel mixing enables it to learn cross-channel correlations. 
In the experiments, we employed three different flavors of the TSMixer model: TSMixer vanilla (referred as TSMixer throughout the text), TSMixer with cross-channel mixing enabled (TSMixer-CM), and TSMixer with cross-channel reconciliation head (TSMixer-CC). We request the readers to refer to \cite{tsmixer} for further details about these variants.

\section{Literature Survey}
\label{appendix:priorarts}
\subsection{Multivariate Time Series Forecasting}
Statistical approaches for time series forecasting, such as SARIMAX and Exponential Smoothing, generally generate forecasts independently for each time series~\cite{hyndman_book}. These methods are essentially univariate and do not build a single model by learning from multiple time series. On the other hand, more advanced models, built upon machine/deep learning techniques, including LightGBM-based models~\cite{makridakis2022m5,hpro}, N-BEATS~\cite{nbeats}, and DeepAR~\cite{deepar}, have the capability to learn from multiple time series. However, these models still follow univariate approaches, thus ignoring any potential cross-channel correlations.

Advanced multivariate forecasting models mostly involve deep neural networks, specifically the transformer~\cite{transformer} architecture.
A series of transformer-based model have been proposed in the last few years including Informer~\cite{informer}, Autoformer~\cite{autoformer}, and FEDFormer~\cite{fedformer}.
Although these models outperformed all the prior arts, the DLinear~\cite{dlinear} model showed that an embarrassingly simple linear model can beat these models by following a few empirically established steps like time series decomposition, normalization, and channel-independent modeling.

PatchTST~\cite{patchtst} showed that transformers can be effective for forecasting if the input time series is patched or segregated in multiple windows, and subsequently, modeled by a transformer.
The patching operation helps preserve local semantic information, accommodates a longer history, and reduces computation time.
The PatchTST model outperformed all prior transformer-based models and the DLinear model.

Although PatchTST reinstated faith in transformers for time series modeling, transformer-based models are generally resource-intensive, with slow execution and a high memory footprint. The recently proposed TSMixer model~\cite{tsmixer} addresses these challenges effectively. TSMixer, built on the MLPMixer architecture~\cite{mlpmixer}, stands out for its exceptional speed and lightweight design. It has attained state-of-the-art (SOTA) performance on benchmark datasets, demonstrating a 2-3X reduction in both execution time and memory usage.

Recently, several new Transformer- and Mixer-based architectures have been proposed.
The iTransformer model~\cite{itransformer} applies attention and MLP modules to the inverted dimension. Instead of operating on the temporal tokens, these operations are applied to the variate tokens, resulting in ``variate-unmixed representations''. This approach is claimed to enhance generalization across different channels and improve the use of arbitrary context lengths.
The TimeMixer model~\cite{timemixer} leverages the observation that time series exhibit unique patterns at different sampling scales. By utilizing different MLPMixer blocks, it aims to capture both microscopic and macroscopic information to produce more accurate forecasts.
Similarly, the recent TimesNet model~\cite{timesnet} disentangles the complex multi-periodicity in a time series into intra-period and inter-period variations. It then learns time series representations using an Inception block, enhancing the model's ability to capture intricate patterns in the data.

\subsection{Pre-trained Models for Time Series}
One major drawback of all the above models is that they need to be trained in-domain. Hence, none of these models can be transferred to out-of-domain data with zero or minimal training. This approach has been found to be extremely beneficial in the natural language processing (NLP) domain with the invention of BERT~\cite{bert} and GPT~\cite{gpt} models.

However, this is an extremely challenging task in the time series domain because of the unavailability of a publicly accessible large pre-training corpora. There are multiple independent time series datasets, but, unlike in NLP, these datasets differ significantly in important characteristics such as the domain of the data (e.g., retail, sensor data, traffic, etc.), the number of channels, temporal resolution, and length. This makes it hard to train a single model on all the datasets together.

Hence, a few prior works have focused on experimenting with \textit{same-dataset} self-supervised learning for time series~\cite{timae,laST,CoST,ts2vec}. These methods learn a time series representation from the \textit{train} split of a dataset, build a forecaster on top of the learned representation on the same data, and then evaluate it on the \textit{test} split of the same dataset. 
Although these approaches have demonstrated promising results, they do not provide evidence of the transfer capability of the model between datasets.

Subsequently, models such as SimMTM~\cite{dong2023simmtm} and TF-C~\cite{tfc} have demonstrated the transfer capabilities of their models between pairs of datasets. These pairs are carefully chosen so that the \textit{source} (the dataset where the model is pre-trained) and \textit{target} (the dataset where the model is fine-tuned and tested) datasets share some matching properties.
For instance, SimMTM showcased its few-shot capability by selecting ETTH2 as the source data and ETTH1 as the target data. Both ETTH1 and ETTH2 are collected from Electricity Transformers at two stations, denoting data from a similar domain.
TF-C demonstrated the transferability of the model across four different (source, target) pairs, such as (ECG, EMG) and (FD-A, FD-B), where domain-similarity exists in both the source and target datasets.

To overcome this limitation, the time series research community is increasingly focused on developing General Pre-Trained (GPT) or Foundation Models (FM) for time-series forecasting, capable of effectively transferring knowledge to new target TS datasets. This growing interest led to the release of several ``large'' and ``massive'' pre-trained time-series models for forecasting in early 2024, generating significant excitement among researchers. Notable releases include Moment~\cite{moment}, TimesFM~\cite{timesfm}, Chronos~\cite{chronos}, Moirai~\cite{moirai}, and Lag-llama~\cite{lagllama}, all of which set strong benchmarks in zero-shot forecasting.
The Moment~\cite{moment} model pre-trains a Transformer encoder model in a univariate way on a collected set of diverse ``Time Series Pile''. Moment is pre-trained with mask reconstruction objective, and it can be fine-tuned on a downstream forecasting task.
The TimesFM~\cite{timesfm} pre-trains a decoder-style attention model (with causal self-attention) in univariate fashion on a large collection of real world and synthetic datasets.
The Chronos~\cite{chronos} model tokenizes the input time series, and feed the tokens into a large langugae model (specifically the T5 model). Chronos is pre-trained in a univariate fashion. During inference, Chronos auto-regressively samples tokens and map them to the numerical values via dequantization. Chronos is trained on a large corpora of time series including synthetic data for better generalization.
The Moirai~\cite{moirai} model pre-trains a Transformer encoder on a massive collection of ``LOTSA'' dataset (27B time points).
Moirai masks the forecast horizon of each target channel and performs mask reconstruction.
The flattening operation of all channels in a multivariate time series enables Moirai to pre-train on ``any-variate'' settings.
The Lag-Llama~\cite{lagllama} model pre-trains a decoder-only Transformer model that utilizs the time series lags as covariates.
Lag-Llama is pre-trained on a large collection of diverse time series datasets in a univariate fashion. All the above models are open-sourced and used in our experiments for comparison. However, closed-source models such as TimeGPT~\cite{timegpt} are not included due to their inaccessibility.

\subsection{Pre-trained LLMs for Time Series}
Parallel to the above trend of general pre-trained TS models, there has been a notable increase in the adoption of pre-trained large language models (LLMs) for time series tasks. These models are approached as cross-domain transfer learning problems. 
The LLMTime model~\cite{llmtime} feeds the time series values as text representations and demonstrates promising performance in a zero-shot setting. The GPT4TS model~\cite{gpt4ts} adopts a pre-trained LLM like GPT and fine-tunes only the input embedding layer, normalization layers, and output layer. Specifically, it does not alter the self-attention weights and feed-forward layers. 
The Time-LLM~\cite{timellm} model proposed a reprogramming framework, where they reuse existing LLMs for forecasting while keeping the LLM backbone intact.
The overall approach to building a pre-trained model for time series from LLMs is promising, but it does not model cross-channel correlations observed in many multivariate time series datasets. Moreover, these LLMs are very large and exhibit slow execution and a large memory footprint.

\section{Datasets}
\label{appendix:data_info}


\begin{table*}[t]
    \centering
    \scriptsize
    \begin{tabular}{|c|c|c|}
        \toprule
        \textbf{Source} & \textbf{Dataset} & \textbf{Resolution} \\\hline
        \multirow{14}{*}{\textbf{Monash}} & kaggle\_web\_traffic\_dataset\_without\_missing\_values & daily \\
        & nn5\_daily\_dataset\_without\_missing\_values & daily \\
        & solar\_10\_minutes\_dataset + \textbf{Downsample} & 10 mins, 30 mins, hourly \\
        & australian\_electricity\_demand\_dataset + \textbf{Downsample} & 30 mins, hourly, daily \\
        & solar\_4\_seconds\_dataset + \textbf{Downsample} & 4 seconds, 10 mins, 15 mins, 30 mins, hourly \\
        & wind\_4\_seconds\_dataset + \textbf{Downsample} & 4 seconds, 10 mins, 15 mins, 30 mins, hourly \\
        & us\_births\_dataset & daily \\
        & saugeenday\_dataset & daily \\ 
        & sunspot\_dataset\_without\_missing\_values & daily \\
        & australian\_weather\_dataset & daily \\
        & kdd\_cup\_2018\_dataset\_without\_missing\_values & hourly \\
        & bitcoin\_dataset\_without\_missing\_values & daily \\
        & wind\_farms\_minutely\_dataset\_without\_missing\_values + \textbf{Downsample} & minutely, 10 mins, 15 mins, 30 mins, hourly \\
        & london\_smart\_meters\_dataset\_without\_missing\_values\ + \textbf{Downsample} & 30 mins, hourly, daily \\ \hline
        \multirow{9}{*}{\textbf{LibCity}} & PEMS03 + \textbf{Downsample} & 5 mins, 10 mins, 15 mins, 30 mins, hourly \\
        & PEMS04 + \textbf{Downsample} & 5 mins, 10 mins, 15 mins, 30 mins, hourly \\
        & PEMS07 + \textbf{Downsample} & 5 mins, 10 mins, 15 mins, 30 mins, hourly \\
        & PEMS08 + \textbf{Downsample} & 5 mins, 10 mins, 15 mins, 30 mins, hourly \\
        & PEMS\_BAY + \textbf{Downsample} & 5 mins, 10 mins, 15 mins, 30 mins, hourly \\
        & LOS\_LOOP + \textbf{Downsample} & 5 mins, 10 mins, 15 mins, 30 mins, hourly \\
        & LOOP\_SEATTLE + \textbf{Downsample} & 5 mins, 10 mins, 15 mins, 30 mins, hourly \\
        & SZ\_TAXI + \textbf{Downsample} & 15 mins, 30 mins \\
        & Q-TRAFFIC + \textbf{Downsample} & 15 mins, 30 mins, hourly \\ \bottomrule
    \end{tabular}
    \caption{List of pre-training datasets. A dataset with ``+ \textbf{Downsample}'' denotes that the proposed Diversity Resolution Sampling (DRS) has been applied on that dataset to generate new diverse datasets at frequencies lower than the original frequency of the data. Please note that, these pre-training datasets have no overlap with the evaluation datasets. Specifically, the australian\_electricity\_demand\_dataset and australian\_weather\_dataset used in pre-training are completely different (\textit{w.r.t} location, measured variables, type, resolution, length, etc.) from the standard Electricity (ECL) and Weather dataset used in the evaluation. Please note that, the last three datasets in the Libcity section have been excluded from the pre-training process for the model releases intended for \href{https://huggingface.co/ibm-granite/granite-timeseries-ttm-r2}{enterprise-use}.}
    \label{tab:pre-train_datasets}
\end{table*}

\subsection{List of Pre-training Datasets}
\label{appendix:pretrain_datasets}
Pre-training employs a subset of $\sim$1B samples from Monash~\cite{monash} and Libcity~\cite{libcity,moirai} data collection, where Monash results in $\sim$250M samples and LibCity accounts for the rest.
In this estimate, one sample denotes a pair of training windows: $X \in \mathbb{R}^{1\times sl}$ and $Y \in \mathbb{R}^{1\times fl}$.
We employ a subset of the datasets available in the Monash forecasting data repository~\cite{monash} available at \url{https://forecastingdata.org/}.
Since our primary focus in this study is long term forecasting with forecast length ranging from 96 to 720, it is not possible to use yearly, monthly, quarterly, or weekly datasets due to their short lengths.
Hence, we skip a few datasets of short lengths.
The Monash datasets used are available under a Creative Commons Attribution 4.0 International license.
For LibCity, we employ all datasets released by the Moirai authors~\cite{moirai}, available at \url{https://huggingface.co/datasets/Salesforce/lotsa_data/tree/main} (except the Rotterdam dataset which was not available during our experimentation). The LibCity datasets at the above link were released under an Apache 2.0 license.
The final list of all pre-training datasets is shown in Table~\ref{tab:pre-train_datasets}. Please note that, the last three datasets in the Libcity section have been excluded from the pre-training process for the model releases intended for \href{https://huggingface.co/ibm-granite/granite-timeseries-ttm-r2}{enterprise-use}. 

\subsubsection{Temporal cross-validation}
Temporal cross-validation is used to chronologically split all the time series into train and validation parts.
During pre-training, moving windowing technique is used to create $\left( \bm{X}, \bm{Y} \right)$ pairs of lengths $sl$ and $fl$ respectively.
Please note that, these pre-training datasets have no overlap with the evaluation datasets. Specifically, the australian\_electricity\_demand\_dataset and australian\_weather\_dataset used in pre-training are completely different (\textit{w.r.t} location, measured variables, type, resolution, length, etc.) from the standard Electricity (ECL) and Weather dataset used in the evaluation. 

\begin{table*}[t]
    \centering
    \begin{tabular}{|c|c|c|c|c|c|c|p{1cm}|}
        \hline
         Set & Dataset & Resolution & Length & \makecell[l]{Total\\\#Channels} & \makecell[l]{\#Target\\variables} & \makecell[l]{\#Exog.\\variables} & Source \\ \hline
        \multirow{7}{*}{\textbf{D1}}& ETTH1 & 1 hour & 17420 & \multirow{4}{*}{7}&  \multirow{4}{*}{7} & \multirow{7}{*}{Not Applicable} & \multirow{4}{*}{\cite{autoformer}} \\ \cline{2-4}
        & ETTH2 & 1 hour & 17420 &  &  & & \\ \cline{2-4}
        & ETTM1 & 15 minutes & 69680 &  & & &  \\ \cline{2-4}
        & ETTM2 & 15 minutes & 69680 &  &  & &  \\ \cline{2-6} \cline{8-8}
        & Weather & 10 minutes & 52696 & 21 & 21 & & \cite{autoformer} \\ \cline{2-6}\cline{8-8}
        & ECL & 1 hour & 26304 & 321 & 321 & & \cite{autoformer} \\ \cline{2-6} \cline{8-8}
        & Traffic & 1 hour & 17544 & 862 & 862 & & \cite{autoformer} \\ \hline
        \multirow{4}{*}{\textbf{D2}} & BS & 1 hour & 17379 & 14 & 3 & 11 & \cite{misc_bike_sharing_dataset_275} \\ \cline{2-8} 
        & CC & 2 minutes & 5409 & 8 & 2 & 5 & \cite{carbon} \\ \cline{2-8}
        & APP & 10 seconds & 8834 & 39  & 4  & 35 & \cite{ITBiz_data} \\ \cline{2-8}
        & SER & 10 seconds & 8835 & 107 & 72 & 35 & \cite{ITBiz_data}  \\ \hline
    \end{tabular}
    \caption{Details of the evaluation datasets.}
    
    \label{tab:eval_data_details}
\end{table*}

\subsection{List of Evaluation Datasets}
\label{appendix:evaluation_datasets}
Table~\ref{tab:eval_data_details} illustrates various characteristics of the eleven evaluation datasets. Below, we present the details.

\textbf{Set D1:}
For zero/few/full-shot evaluation, we utilize seven multivariate time series datasets that have consistently been employed in the literature. Below, we offer a brief overview of these datasets.

\begin{enumerate}
    \item \textbf{ETT datasets:} The four ETT datasets~\cite{informer} (ETTH1, ETTH2, ETTM1, ETTM2) contain multivariate time series data collected from electrical transformers at two stations. ETTH1 and ETTH2 are collected at an hourly interval, while ETTM1 and ETTM2 are collected every 15 minutes. All four datasets have 7 channels.

    \item \textbf{Weather:} The weather dataset consists of 21 channels, which serve as weather indicators. It is collected at 10-minute intervals at the Max Planck Institute of Biogeochemistry weather station.

    \item \textbf{Electricity (ECL):} The Electricity dataset, also known as the ECL dataset, comprises the hourly electricity consumption data of 321 clients.

    \item \textbf{Traffic:} This dataset records the hourly rates of road occupancy on the San Francisco Freeways using 862 sensors.
\end{enumerate}
We used the datasets provided in the repository of the Autoformer paper~\cite{autoformer}\footnote{Available at \url{https://github.com/thuml/Autoformer}}.
For all the D1 datasets, we execute the same train/validation/test splitting as was performed in the literature~\cite{informer,autoformer,patchtst,tsmixer}.

\textbf{Set D2:}
To assess the effectiveness of the proposed TTM model in extracting information from exogenous channels, we conduct evaluations on four additional datasets that are known to contain exogenous or control variables.

\begin{enumerate}
    \item \textbf{Bike Sharing (BS):} The Bike Sharing dataset~\cite{misc_bike_sharing_dataset_275} documents the hourly rental counts of bikes from the Capital Bikeshare system in Washington D.C., USA, spanning the years 2011 to 2012. Rental counts are typically associated with environmental and seasonal conditions. Consequently, this 14-channel dataset encompasses various weather-related features. Our goal was to forecast all three rental counts: ``casual'', ``registered'', and ``cnt'' (total count). As the remaining 11 features are consistently available at all future time points, they are treated as exogenous variables in our experiment.
    
    \item \textbf{Carbon Capture Plant (CC):} The Carbon Capture Plant data~\cite{carbon} records the emission profiles of ``2-amino-2-methyl-1-propanol''~(AMP) and ``piperazine''~(Pz) collected at every 2 minutes interval. We utilize the 8-channel dataset made available in the official repository~\cite{carbon}. Among the remaining 6 channels, the following 5 serve as control variables: [``TI-19'',``FI-19'', ``TI-3'', ``FI-11'', ``TI-1213'']. The remaining 1 variable is treated as a conditional variable (as it is neither a target variable nor available during the forecast period to consider it as exogenous). For additional details, please refer to the supplementary materials of \cite{carbon}.

    \item \textbf{Service (SER):} This dataset pertains to the cloud-based ``Stan's Robot Shop'' application, managed by Instana. It simulates a user's e-commerce experience, encompassing site access to shipping, utilizing a load generator. Intermittent fault injection introduces diverse IT events. The dataset provides business KPIs for services (e.g., payment, catalog) and IT events tracked by Instana. Sampling occurs every 10 seconds due to high traffic and event frequency. For our experiments, all business KPIs are treated as target variables and IT events are treated as exogenous variables and the goal of our forecasting is to predict the business KPIs given the IT events.

    \item \textbf{Application (APP):} This dataset is similar to the SER data, but it captures KPIs for the entire application instead of capturing at the service level. Even in this case, all business KPIs are treated as target variables and IT events are treated as exogenous variables and the goal of our forecasting is to predict the business KPIs given the IT events.
\end{enumerate}

\section{TTM Model Hyper-parameters and Baselines}
\label{app:sec:TTM Model Hyper-parameters and Baselines}
\subsection{Pretraining}
Pre-training is performed in a distributed fashion with 50 CPUs and 6 NVIDIA A100 GPUs. 
Standard model configurations are as follows: patch length $pl$ = 64 (when $sl$ is 512), 128 (when $sl$ is 1024 or 1536) and 8 (when $sl$ is 96); stride $s$ = $pl$ (\ie non-overlapping patches), number of patches $n=sl/pl$, number of levels in backbone $L$ = 3, number of TTM blocks per level $M$ = 2, number of decoder layers = 2, batch size $b$ = 4500, number of epochs $ep$ = 20, and dropout $do$ = 0.4. PatchTSMixer-specific hyperparameters include feature scaler $fs$ = 3, hidden feature size $hf$ = $fs*pl$, expansion feature size $ef = hf*2$. Please note that $hf$ and $n$ will change across TTM blocks based on the adaptive patching strategy. Resolution prefix tuning is enabled by default on all variants other than \ttmq. Decoder channel-mixing and exogenous mixer blocks are disabled during pre-training and enabled during fine-tuning based on the dataset requirement.

\subsection{Fine-tuning} Most model parameters remain the same from pretraining except the following parameters. Head dropout is changed during finetuning based on the target dataset used (0.7 for smaller ETT datasets and 0.2 for other datasets). Likewise, the batch size is set to 8 for Traffic, 32 for Electricity, and 64 for all other datasets. Moreover, decoder channel-mixing and exogenous mixer block are enabled for datasets that need cross-channel modelling (i.e. D2 datasets). Unlike pre-training, fine-tuning is executed in just 1 A100 GPU as it is a fast process. All these hyper-parameters are selected based on the validation performance, and the final test results are reported in the paper. 

\subsection{Computational Benefits of TTM over existing models - Setup details}
\label{appendix:ttm_comp_2}

Table ~\ref{tab:n_timing} compares the computational benefits of TTM over existing TS-pretrained models and reports the following metrics: (i) GPU Inference Time per batch (in milliseconds (ms)), (ii) CPU Inference Time per batch (in seconds (s)), (iii) Max GPU Memory used during inference (in GB), (iv) Params: Total parameters of the models (in Millions). Experiments are conducted using $sl$ = 512, $fl$ = 96, and  batch size = 32 in one A100 80GB GPU, 16 cores with 256GB memory. GPU is not enabled while capturing the CPU time.  Since many pre-trained models process data in a purely univariate fashion, while TTM processes data in a multi-variate fashion, we set the number of channels $c$ to 1 for this experiment so that, the number of samples per batch remains the same across all models for a fair comparison.  In addition, we used a small batch size of 32 for this experiment, as many pre-trained models (like \chronusl) were encountering out-of-memory (OOM) errors with high batch sizes. 
For this experiment, we set the number of probabilistic samples to 1 (i.e., $\texttt{num\_samples} = 1$) for probabilistic algorithms (such as Chronos or Lag-Llama) to compute their fastest possible runtime.
Note, that for forecast accuracy comparison, we set the number of samples to 100 for Lag-Llama and 20 for Chronos as suggested in their open-source code examples.
All the baselines algorithms were evaluated using their open-sourced inference APIs as detailed in Section~\ref{appendix:baselines}.
Please note that the computational  benefits of TTM will further amplify if we use higher batch sizes or high number of channels as our models are extremely small and can process multiple channels at the same time using the channel-independence approach~\cite{patchtst}. 



\subsection{Baseline Implementation Details}
\label{appendix:baselines}
We report the implementation details for all the baselines in Table~\ref{tab:baseline-implementations}.
\begin{table*}[t]
    \centering
    \scriptsize
    \begin{tabular}{|l|l|l|l|p{3cm}|}
    \toprule
        \textbf{Category} & \textbf{Baseline} & \textbf{Used in Table}  & \makecell[l]{\textbf{Results}\\\textbf{Generated From}} & \textbf{\makecell{Link to the\\used implementation}} \\ \midrule
        
        \multirow{7}{*}{\makecell[l]{(a) TS pre-trained\\SOTA models}} & Moirai~\cite{moirai} & \makecell[l]{Zero-shot in Table~\ref{tab:n_zs_moirai_avg} and \ref{tab:n_zs_moirai_full}} & \makecell[l]{Table 6 and 22 of \cite{moirai}} & N/A \\ \cline{2-5}
        
        & Moirai~\cite{moirai} & \makecell[l]{Runtime in Table~\ref{tab:n_timing},\\Exog. expt. in Table~\ref{tab:exog_mixer_avg}} & \makecell[l]{Official implementation} & \href{https://github.com/SalesforceAIResearch/uni2ts}{\texttt{uni2ts}} \\ \cline{2-5}
        
        & TimesFM~\cite{timesfm} & \makecell[l]{Zero-shot in Table~\ref{tab:n_zs_moirai_avg} and \ref{tab:n_zs_moirai_full},\\Runtime in Table~\ref{tab:n_timing},\\Exog. expt. in Table~\ref{tab:exog_mixer_avg}} & \makecell[l]{Official implementation} & \href{https://github.com/google-research/timesfm}{\texttt{timesfm}} \\ \cline{2-5}
        
        & Chronos~\cite{chronos} & \makecell[l]{Zero-shot in Table~\ref{tab:n_zs_2_avg} and \ref{tab:n_zs_2_full},\\Runtime in Table~\ref{tab:n_timing}} & \makecell[l]{Official implementation} & \href{https://github.com/amazon-science/chronos-forecasting}{\texttt{chronos-forecasting}} \\ \cline{2-5}
        
        & Lag-Llama~\cite{lagllama} & \makecell[l]{Zero-shot in Table~\ref{tab:n_zs_2_avg} and \ref{tab:n_zs_2_full},\\Runtime in Table~\ref{tab:n_timing}} & \makecell[l]{Official implementation} & \href{https://github.com/time-series-foundation-models/lag-llama}{\texttt{lag-llama}} \\ \cline{2-5}
        
        & Moment~\cite{moment} & \makecell[l]{Full-shot Head-probing\\in Table~\ref{tab:n_hp_avg} and \ref{tab:n_hp_full}} & \makecell[l]{Table 2 of \cite{moment}} & N/A \\ \cline{2-5}
        
        & Moment~\cite{moment} & \makecell[l]{Runtime in Table~\ref{tab:n_timing}} & \makecell[l]{Official Implementation} & \href{https://github.com/moment-timeseries-foundation-model/moment}{\texttt{moment}} \\ 
        \midrule
        
        \multirow{7}{*}{\makecell[l]{(b) LLM-based TS\\pre-trained models}} & GPT4TS~\cite{gpt4ts} & \makecell[l]{Few-shot 5\% in Table~\ref{tab:n_fs5_avg} and \ref{tab:n_t3_fs5_full}} & \makecell[l]{Table 12 of \cite{gpt4ts}} & N/A \\ \cline{2-5}
        
        & GPT4TS~\cite{gpt4ts} & \makecell[l]{Runtime in Table~\ref{tab:n_timing},\\Exog. expt. in Table~\ref{tab:exog_mixer_avg}} & \makecell[l]{Official Implementation} & \href{https://github.com/DAMO-DI-ML/NeurIPS2023-One-Fits-All}{\texttt{One-Fits-All}} \\ \cline{2-5}
        
        
        & GPT4TS~\cite{gpt4ts} & \makecell[l]{Full-shot Head-probing\\in Table~\ref{tab:n_hp_avg} and \ref{tab:n_hp_full}} & \makecell[l]{Table 2 of \cite{moment}} & N/A\\ \cline{2-5}
        
        & LLMTime~\cite{llmtime} & Zero-shot in Table~\ref{tab:n_zs_llmtime_full} & \makecell[l]{Results available in\\the LLMTime Repository} & \href{https://github.com/ngruver/llmtime}{\texttt{llmtime}} \\ \cline{2-5}
        
        & Time-LLM~\cite{timellm} & \makecell[l]{Full-shot Head-probing\\in Table~\ref{tab:n_hp_avg} and \ref{tab:n_hp_full}} & \makecell[l]{Table 2 of \cite{moment}} & N/A \\ \cline{2-5}
        
        & UniTime~\cite{unitime} & \makecell[l]{Zero-shot in Table~\ref{tab:unitime}} & \makecell[l]{Table 5 of~\cite{unitime}} & N/A \\ \hline
        
        \multirow{7}{*}{\makecell[l]{(b) Self-\\supervised\\pre-trained\\models}} & SimMTM~\cite{dong2023simmtm} & \multirow{7}{*}{Table~\ref{tab:simmtm_2}} & \multirow{7}{*}{\makecell[l]{Directly reported\\from SimMTM paper~\cite{dong2023simmtm}}} & \multirow{7}{*}{N/A} \\  
         & Ti-MAE~\cite{timae} & ~ & ~ &  \\ 
         & TST~\cite{tst} & ~ & ~ &  \\ 
         & LaST~\cite{laST} & ~ & ~ &  \\ 
         & TF-C~\cite{tfc} & ~ & ~ &  \\ 
         & CoST~\cite{CoST} & ~ & ~ &  \\ 
         & TS2Vec~\cite{ts2vec} & ~ & ~ & ~ \\ \hline
         
        \multirow{6}{*}{\makecell[l]{(d) Other SOTA\\Architectures}} & PatchTST~\cite{patchtst} & \multirow{6}{*}{\makecell[l]{Few-shot 5\%in Table~\ref{tab:n_fs5_avg} and \ref{tab:n_t3_fs5_full}}} & Taken from \cite{gpt4ts} & N/A \\ \cline{2-2} \cline{4-5}
        
        & TSMixer~\cite{tsmixer} &  & \makecell[l]{Official Implementation} & \href{https://huggingface.co/docs/transformers/en/model_doc/patchtsmixer}{\texttt{PatchTSMixer}} \\ \cline{2-2}\cline{4-5}

        & TimeMixer~\cite{timemixer} &  & \makecell[l]{Official Implementation} & \makecell[l]{\href{https://github.com/thuml/Time-Series-Library}{\texttt{Time-Series-Library}}} \\ \cline{2-2}\cline{4-5}

        & iTransformer~\cite{itransformer} &  & \makecell[l]{Official Implementation} & \makecell[l]{\href{https://github.com/thuml/Time-Series-Library}{\texttt{Time-Series-Library}}} \\ \cline{2-2}\cline{4-5}

        & TimesNet~\cite{timesnet} &  & \makecell[l]{Taken from \cite{gpt4ts}} & N/A \\\cline{2-2}\cline{4-5}
        & Dinear~\cite{dlinear} &  & \makecell[l]{Taken from \cite{gpt4ts}} & N/A \\\cline{2-5}
        
        & FEDFormer~\cite{fedformer} & \multirow{3}{*}{Full-shot end2end Table~\ref{tab:n_hp_full}} & \multirow{3}{*}{Taken from \cite{moment}} & \multirow{3}{*}{N/A} \\ 
        & Autoformer~\cite{autoformer} &  &  &  \\ 
        & Informer~\cite{informer} &  &  &  \\ 
        \hline

    \end{tabular}
    \caption{Implementation details for the baseline algorithms.}
    \label{tab:baseline-implementations}
\end{table*}

\begin{figure*}[h]
  \begin{minipage}{0.95\textwidth}
    \centering
    \scalebox{0.85}{
    \begin{tabular}{|c|c|c|c|c|c|c|} \hline 
 & \makecell{FL} & \makecell{\ttmq~} & \makecell{\ttms~} & \makecell{\ttmm~} & \makecell{\ttml~}\\ \hline 
\multirow{4}*{\rotatebox{90}{ETTH1}} & 96 & 0.365 & 0.364 & \uline{0.363} & \textbf{0.359} \\ 
 & 192 & 0.393 & \textbf{0.386} & 0.393 & \uline{0.389} \\ 
 & 336 & 0.415 & \textbf{0.404} & 0.406 & \uline{0.405} \\ 
 & 720 & 0.538 & \textbf{0.424} & 0.452 & \uline{0.448} \\ 
\hline 
\multirow{4}*{\rotatebox{90}{ETTH2}} & 96 & 0.285 & 0.277 & \uline{0.271} & \textbf{0.264} \\ 
 & 192 & 0.341 & 0.334 & \uline{0.324} & \textbf{0.321} \\ 
 & 336 & 0.383 & 0.362 & \uline{0.357} & \textbf{0.351} \\ 
 & 720 & 0.441 & 0.408 & \textbf{0.388} & \uline{0.395} \\ 
\hline 
\multirow{4}*{\rotatebox{90}{ETTM1}} & 96 & 0.413 & \uline{0.322} & 0.327 & \textbf{0.318} \\ 
 & 192 & 0.476 & \uline{0.376} & 0.377 & \textbf{0.354} \\ 
 & 336 & 0.553 & 0.407 & \uline{0.395} & \textbf{0.376} \\ 
 & 720 & 0.737 & 0.439 & \uline{0.419} & \textbf{0.398} \\ 
\hline 
\multirow{4}*{\rotatebox{90}{ETTM2}} & 96 & 0.187 & \uline{0.171} & 0.178 & \textbf{0.169} \\ 
 & 192 & 0.261 & \uline{0.238} & \uline{0.238} & \textbf{0.223} \\ 
 & 336 & 0.323 & 0.304 & \uline{0.29} & \textbf{0.276} \\ 
 & 720 & 0.436 & 0.41 & \uline{0.379} & \textbf{0.342} \\ 
\hline 
\multirow{4}*{\rotatebox{90}{Weather}} & 96 & \textbf{0.154} & \uline{0.158} & 0.166 & 0.159 \\ 
 & 192 & \textbf{0.203} & \uline{0.206} & 0.214 & \textbf{0.203} \\ 
 & 336 & 0.256 & 0.256 & \uline{0.254} & \textbf{0.247} \\ 
 & 720 & 0.329 & 0.328 & \uline{0.319} & \textbf{0.314} \\ 
\hline 
\multirow{4}*{\rotatebox{90}{Electricity}} & 96 & 0.169 & 0.166 & \uline{0.157} & \textbf{0.152} \\ 
 & 192 & 0.196 & 0.191 & \textbf{0.174} & \uline{0.179} \\ 
 & 336 & 0.209 & 0.207 & \uline{0.195} & \textbf{0.193} \\ 
 & 720 & 0.264 & 0.255 & \uline{0.25} & \textbf{0.243} \\ 
\hline 
\multirow{4}*{\rotatebox{90}{Traffic}} & 96 & 0.518 & 0.514 & \uline{0.476} & \textbf{0.462} \\ 
 & 192 & 0.548 & 0.544 & \uline{0.5} & \textbf{0.491} \\ 
 & 336 & 0.55 & 0.575 & \uline{0.51} & \textbf{0.509} \\ 
 & 720 & 0.605 & 0.622 & \uline{0.571} & \textbf{0.547} \\ 
\hline 
\multicolumn{2}{|c|}{\makecell{\textbf{Model Size}}} & \textbf{1M} & \textbf{1M} & \textbf{4M} & \textbf{5M}\\ \hline 
 \end{tabular}

    }
    \captionof{table}{Zero-shot results of all TTM variants on D1 data benchmark across all sliding test windows (standard test protocol).}
    \label{tab:appendix_zs_master}
  \end{minipage}%
\end{figure*}

\begin{figure*}[h]
   \begin{minipage}{0.95\textwidth}
    \centering
    \scalebox{0.85}{
   
\begin{tabular}{|c|c|c|c|c|c|c|} \hline 
 & \makecell{FL} & \makecell{\ttmq~} & \makecell{\ttms~} & \makecell{\ttmm~} & \makecell{\ttml~}\\ \hline 
\multirow{4}*{\rotatebox{90}{ETTH1}} & 96 & 0.366 & 0.364 & \uline{0.363} & \textbf{0.359} \\ 
 & 192 & \uline{0.391} & \textbf{0.387} & 0.393 & 0.394 \\ 
 & 336 & 0.421 & \uline{0.399} & \textbf{0.398} & 0.406 \\ 
 & 720 & - & - & - & - \\ 
\hline 
\multirow{4}*{\rotatebox{90}{ETTH2}} & 96 & 0.282 & 0.277 & \uline{0.271} & \textbf{0.267} \\ 
 & 192 & 0.338 & 0.334 & \uline{0.325} & \textbf{0.321} \\ 
 & 336 & 0.383 & 0.361 & \uline{0.357} & \textbf{0.354} \\ 
 & 720 & - & - & - & - \\ 
\hline 
\multirow{4}*{\rotatebox{90}{ETTM1}} & 96 & 0.359 & \textbf{0.313} & 0.326 & \uline{0.317} \\ 
 & 192 & 0.402 & \uline{0.357} & 0.371 & \textbf{0.355} \\ 
 & 336 & 0.424 & \uline{0.395} & 0.396 & \textbf{0.374} \\ 
 & 720 & 0.575 & 0.437 & \uline{0.418} & \textbf{0.397} \\ 
\hline 
\multirow{4}*{\rotatebox{90}{ETTM2}} & 96 & 0.174 & \uline{0.171} & 0.178 & \textbf{0.17} \\ 
 & 192 & 0.24 & \uline{0.23} & 0.237 & \textbf{0.222} \\ 
 & 336 & 0.299 & 0.293 & \uline{0.284} & \textbf{0.274} \\ 
 & 720 & 0.407 & 0.393 & \uline{0.373} & \textbf{0.345} \\ 
\hline 
\multirow{4}*{\rotatebox{90}{Weather}} & 96 & \textbf{0.152} & \uline{0.154} & 0.162 & 0.155 \\ 
 & 192 & \textbf{0.198} & 0.203 & 0.215 & \uline{0.201} \\ 
 & 336 & \uline{0.25} & 0.252 & 0.262 & \textbf{0.244} \\ 
 & 720 & 0.326 & 0.327 & \uline{0.319} & \textbf{0.316} \\ 
\hline 
\multirow{4}*{\rotatebox{90}{Electricity}} & 96 & \uline{0.142} & 0.146 & 0.15 & \textbf{0.141} \\ 
 & 192 & \uline{0.162} & 0.164 & 0.171 & \textbf{0.16} \\ 
 & 336 & \uline{0.184} & 0.185 & 0.202 & \textbf{0.179} \\ 
 & 720 & \textbf{0.228} & \uline{0.236} & 0.303 & 0.24 \\ 
\hline 
\multirow{4}*{\rotatebox{90}{Traffic}} & 96 & \textbf{0.401} & \uline{0.411} & \uline{0.411} & 0.469 \\ 
 & 192 & \uline{0.425} & \textbf{0.42} & 0.44 & 0.488 \\ 
 & 336 & \textbf{0.437} & 0.468 & \uline{0.46} & 0.512 \\ 
 & 720 & - & - & - & - \\ 
\hline 
\multicolumn{2}{|c|}{\makecell{\textbf{Model Size}}} & \textbf{1M} & \textbf{1M} & \textbf{4M} & \textbf{5M}\\ \hline 
 \end{tabular}
    }
    \captionof{table}{Few-shot 5\% results of all TTM variants on D1 data benchmark across all sliding test windows (standard test protocol).}
    \label{tab:appendix_fs5_master}
  \end{minipage}%
\end{figure*}


\begin{figure*}[h]
  \begin{minipage}{0.95\textwidth}
    \centering
    \scalebox{1.0}{
    
\begin{tabular}{|c|c|c|c|c|c|c|c|c|c|} \hline 
 & \makecell{FL} & \makecell{\ttms~} & \makecell{\ttmm~} & \makecell{\ttml~} & \makecell{\moirais~} & \makecell{\moiraib~} & \makecell{\moirail~} & \makecell{TimesFM}\\ \hline 
\multirow{4}*{\rotatebox{90}{ETTH1}} & 96 & 0.364 & \uline{0.363} & \textbf{0.359} & 0.375 & 0.384 & 0.38 & 0.421 \\ 
 & 192 & \textbf{0.386} & 0.393 & \uline{0.389} & 0.399 & 0.425 & 0.44 & 0.472 \\ 
 & 336 & \textbf{0.404} & 0.406 & \uline{0.405} & 0.412 & 0.456 & 0.514 & 0.51 \\ 
 & 720 & \uline{0.424} & 0.452 & 0.448 & \textbf{0.413} & 0.47 & 0.705 & 0.514 \\ 
\hline 
\multirow{4}*{\rotatebox{90}{ETTH2}} & 96 & 0.277 & \uline{0.271} & \textbf{0.264} & 0.281 & 0.277 & 0.287 & 0.326 \\ 
 & 192 & 0.334 & \uline{0.324} & \textbf{0.321} & 0.34 & 0.34 & 0.347 & 0.4 \\ 
 & 336 & 0.362 & \uline{0.357} & \textbf{0.351} & 0.362 & 0.371 & 0.377 & 0.434 \\ 
 & 720 & 0.408 & \uline{0.388} & 0.395 & \textbf{0.38} & 0.394 & 0.404 & 0.451 \\ 
\hline 
\multirow{4}*{\rotatebox{90}{ETTM1}} & 96 & \uline{0.322} & 0.327 & \textbf{0.318} & 0.404 & 0.335 & 0.353 & 0.357 \\ 
 & 192 & 0.376 & 0.377 & \textbf{0.354} & 0.435 & \uline{0.366} & 0.376 & 0.411 \\ 
 & 336 & 0.407 & 0.395 & \textbf{0.376} & 0.462 & \uline{0.391} & 0.399 & 0.441 \\ 
 & 720 & 0.439 & \uline{0.419} & \textbf{0.398} & 0.49 & 0.434 & 0.432 & 0.507 \\ 
\hline 
\multirow{4}*{\rotatebox{90}{ETTM2}} & 96 & \uline{0.171} & 0.178 & \textbf{0.169} & 0.205 & 0.195 & 0.189 & 0.205 \\ 
 & 192 & \uline{0.238} & \uline{0.238} & \textbf{0.223} & 0.261 & 0.247 & 0.247 & 0.293 \\ 
 & 336 & 0.304 & \uline{0.29} & \textbf{0.276} & 0.319 & 0.291 & 0.295 & 0.366 \\ 
 & 720 & 0.41 & 0.379 & \textbf{0.342} & 0.415 & \uline{0.355} & 0.372 & 0.472 \\ 
\hline 
\multirow{4}*{\rotatebox{90}{Weather}} & 96 & \textbf{0.158} & 0.166 & \uline{0.159} & 0.173 & 0.167 & 0.177 & - \\ 
 & 192 & \uline{0.206} & 0.214 & \textbf{0.203} & 0.216 & 0.209 & 0.219 & - \\ 
 & 336 & 0.256 & \uline{0.254} & \textbf{0.247} & 0.26 & 0.256 & 0.277 & - \\ 
 & 720 & 0.328 & \uline{0.319} & \textbf{0.314} & 0.32 & 0.321 & 0.365 & - \\ 
\hline 
\multirow{4}*{\rotatebox{90}{Electricity}} & 96 & 0.166 & \uline{0.157} & \textbf{0.152} & 0.205 & 0.158 & \textbf{0.152} & - \\ 
 & 192 & 0.191 & \uline{0.174} & 0.179 & 0.22 & \uline{0.174} & \textbf{0.171} & - \\ 
 & 336 & 0.207 & 0.195 & 0.193 & 0.236 & \textbf{0.191} & \uline{0.192} & - \\ 
 & 720 & 0.255 & 0.25 & 0.243 & 0.27 & \textbf{0.229} & \uline{0.236} & - \\ 
\hline 
 \end{tabular}
    }
    \captionof{table}{Zero-shot results of TTM over Moirai (ICML'24) and TimesFM (ICML'24). Electricity and Weather results for TimesFM are not reported as they were used by TimesFM for pretraining. Similarly, Traffic data was used in both Moirai and TimesFM pre-training, hence, skipped in this comparison.}
    \label{tab:n_zs_moirai_full}
  \end{minipage}%
\end{figure*}

\begin{figure*}[h]
  \begin{minipage}{0.95\textwidth}
    \centering
    \setlength{\tabcolsep}{2pt}
    \scalebox{.6}{

  \begin{tabular}{|c|c|c|c|c|c|c|c|c|c|c|c|c|c|c|c|} \hline 
  & \multicolumn{7}{|c|}{\textbf{Zero-shot from Pre-Trained Models}} & \multicolumn{8}{|c|}{\textbf{Full-shot End2End Training with short context length setting}} \\ \hline
 \makecell{Data} & \makecell{\ttms~} & \makecell{\ttmm~} & \makecell{\ttml~} & \makecell{\moirais~} & \makecell{\moiraib~} & \makecell{\moirail~} & \makecell{TimesFM} & \makecell{iTransformer} & \makecell{TimesNet} & \makecell{PatchTST} & \makecell{Crossformer} & \makecell{TiDE} & \makecell{Dlinear} & \makecell{SCINet} & \makecell{FEDFormer}\\ \hline 
ETTH1 & \textbf{0.394} & 0.404 & \uline{0.4} & \uline{0.4} & 0.434 & 0.51 & 0.479 & 0.454 & 0.458 & 0.469 & 0.529 & 0.541 & 0.456 & 0.747 & 0.44 \\ 
ETTH2 & 0.345 & \uline{0.335} & \textbf{0.333} & 0.341 & 0.346 & 0.354 & 0.403 & 0.383 & 0.414 & 0.387 & 0.942 & 0.611 & 0.559 & 0.954 & 0.437 \\ 
ETTM1 & 0.386 & \uline{0.38} & \textbf{0.362} & 0.448 & 0.382 & 0.39 & 0.429 & 0.407 & 0.4 & 0.387 & 0.513 & 0.419 & 0.403 & 0.486 & 0.448 \\ 
ETTM2 & 0.281 & \uline{0.271} & \textbf{0.252} & 0.3 & 0.272 & 0.276 & 0.334 & 0.288 & 0.291 & 0.281 & 0.757 & 0.358 & 0.35 & 0.571 & 0.305 \\ 
Weather & \uline{0.237} & 0.238 & \textbf{0.231} & 0.242 & 0.238 & 0.26 & - & 0.258 & 0.259 & 0.259 & 0.259 & 0.271 & 0.265 & 0.292 & 0.309 \\ 
Electricity & 0.205 & 0.194 & 0.192 & 0.233 & \uline{0.188} & \uline{0.188} & - & \textbf{0.178} & 0.193 & 0.216 & 0.244 & 0.252 & 0.212 & 0.268 & 0.214 \\ 
\hline 
\multicolumn{4}{|c|}{\makecell{\textbf{\ttms~} \textit{f-imp(\%)}}} & \textbf{6\% $\uparrow$ } & \textbf{1\% $\downarrow$ } & \textbf{4\% $\uparrow$ } & \textbf{15\% $\uparrow$ } & \textbf{4\% $\uparrow$ } & \textbf{7\% $\uparrow$ } & \textbf{7\% $\uparrow$ } & \textbf{34\% $\uparrow$ } & \textbf{22\% $\uparrow$ } & \textbf{15\% $\uparrow$ } & \textbf{37\% $\uparrow$ } & \textbf{13\% $\uparrow$ } \\ 
\multicolumn{4}{|c|}{\makecell{\textbf{\ttmm~} \textit{f-imp(\%)}}} & \textbf{7\% $\uparrow$ } & \textbf{1\% $\uparrow$ } & \textbf{6\% $\uparrow$ } & \textbf{16\% $\uparrow$ } & \textbf{6\% $\uparrow$ } & \textbf{8\% $\uparrow$ } & \textbf{8\% $\uparrow$ } & \textbf{34\% $\uparrow$ } & \textbf{23\% $\uparrow$ } & \textbf{16\% $\uparrow$ } & \textbf{39\% $\uparrow$ } & \textbf{15\% $\uparrow$ } \\ 
\multicolumn{4}{|c|}{\makecell{\textbf{\ttml~} \textit{f-imp(\%)}}} & \textbf{10\% $\uparrow$ } & \textbf{4\% $\uparrow$ } & \textbf{9\% $\uparrow$ } & \textbf{19\% $\uparrow$ } & \textbf{9\% $\uparrow$ } & \textbf{11\% $\uparrow$ } & \textbf{11\% $\uparrow$ } & \textbf{36\% $\uparrow$ } & \textbf{26\% $\uparrow$ } & \textbf{19\% $\uparrow$ } & \textbf{40\% $\uparrow$ } & \textbf{17\% $\uparrow$ } \\ 
\hline 
 \end{tabular}
 
     }
    \captionof{table}{\scriptsize{Zero-shot Forecast-Improvement \textit{(f-imp)} of TTM over Moirai (ICML'24), TimesFM (ICML'24) and other popular architectures (full shot trained with context length 96).
    MSE averaged across Fls: $\{96,192,336,720\}$. Electricity and Weather results for TimesFM are not reported as they were used by TimesFM for pretraining. Similarly, Traffic data was used in both Moirai and TimesFM pre-training. Full-shot and Moirai results reported from~\cite{moirai}, TimesFM results were generated from their released code.}}
    \label{tab:n_zs_moirai_fullshot_avg}
  \end{minipage}%
\end{figure*}


\begin{figure*}[h]
  \begin{minipage}{0.95\textwidth}
    \centering
    \scalebox{.78}{

 \begin{tabular}{|c|c|c|c|c|c|c|c|c|c|c|} \hline 
 & \makecell{FL} & \makecell{\ttms~} & \makecell{\ttmm~} & \makecell{\ttml~} & \makecell{\chronust~} & \makecell{\chronuss~} & \makecell{\chronusb~} & \makecell{\chronusl~} & \makecell{Lag-llama}\\ \hline 
\multirow{5}*{\rotatebox{90}{ETTH1}} & 24 & 0.195 & 0.243 & 0.217 & 0.207 & 0.217 & \uline{0.163} & \textbf{0.161} & 0.372 \\ 
 & 48 & \textbf{0.193} & 0.244 & \uline{0.226} & 0.33 & 0.277 & 0.228 & \uline{0.226} & 0.376 \\ 
 & 60 & \uline{0.209} & 0.212 & \textbf{0.2} & 0.295 & 0.331 & 0.303 & 0.311 & 0.321 \\ 
 & 96 & \textbf{0.194} & 0.209 & \uline{0.207} & 0.424 & 0.383 & 0.272 & 0.298 & 0.299 \\ 
 & 192 & 0.231 & \uline{0.227} & \textbf{0.221} & 0.299 & 0.303 & 0.295 & 0.335 & 0.303 \\ 
\hline 
\multirow{5}*{\rotatebox{90}{ETTH2}} & 24 & 0.12 & 0.149 & 0.158 & 0.135 & 0.12 & \uline{0.086} & \textbf{0.065} & 0.216 \\ 
 & 48 & \uline{0.151} & 0.18 & 0.197 & 0.186 & 0.207 & 0.194 & \textbf{0.139} & 0.187 \\ 
 & 60 & \textbf{0.166} & \uline{0.175} & 0.193 & 0.217 & 0.187 & 0.195 & 0.2 & 0.182 \\ 
 & 96 & \textbf{0.111} & \uline{0.13} & 0.155 & 0.197 & 0.156 & 0.22 & 0.207 & 0.135 \\ 
 & 192 & \textbf{0.105} & 0.123 & \uline{0.107} & 0.151 & 0.131 & 0.125 & 0.162 & 0.122 \\ 
\hline 
\multirow{5}*{\rotatebox{90}{ETTM1}} & 24 & \textbf{0.141} & 0.247 & \uline{0.187} & 0.285 & 0.196 & 0.284 & 0.247 & 0.207 \\ 
 & 48 & \textbf{0.161} & 0.241 & \uline{0.182} & 0.704 & 0.343 & 0.207 & 0.311 & 0.881 \\ 
 & 60 & \textbf{0.153} & 0.205 & \uline{0.166} & 0.959 & 0.545 & 0.536 & 0.818 & 0.958 \\ 
 & 96 & \textbf{0.172} & 0.209 & \uline{0.184} & 0.854 & 0.589 & 0.508 & 0.538 & 1.006 \\ 
 & 192 & 0.402 & \uline{0.294} & \textbf{0.232} & 1.394 & 0.757 & 0.914 & 0.774 & 1.16 \\ 
\hline 
\multirow{5}*{\rotatebox{90}{ETTM2}} & 24 & \textbf{0.04} & 0.063 & 0.068 & \textbf{0.04} & \uline{0.041} & 0.054 & 0.075 & 0.325 \\ 
 & 48 & \textbf{0.144} & 0.16 & \uline{0.152} & 0.379 & 0.253 & 0.236 & 0.198 & 0.337 \\ 
 & 60 & \uline{0.125} & 0.143 & \textbf{0.119} & 0.274 & 0.233 & 0.222 & 0.222 & 0.294 \\ 
 & 96 & 0.137 & \uline{0.108} & \textbf{0.101} & 0.147 & 0.12 & 0.171 & 0.192 & 0.314 \\ 
 & 192 & 0.175 & \uline{0.165} & \textbf{0.146} & 0.191 & 0.223 & 0.265 & 0.246 & 0.268 \\ 
\hline 
\multirow{5}*{\rotatebox{90}{Weather}} & 24 & 0.015 & 0.015 & 0.016 & \textbf{0.008} & 0.011 & \uline{0.009} & 0.01 & 0.084 \\ 
 & 48 & 0.018 & \uline{0.017} & 0.026 & 0.024 & \uline{0.017} & \textbf{0.016} & 0.022 & 0.105 \\ 
 & 60 & \textbf{0.024} & 0.03 & 0.054 & 0.036 & 0.04 & 0.029 & \uline{0.028} & 0.141 \\ 
 & 96 & \uline{0.034} & \textbf{0.025} & 0.053 & 0.042 & 0.068 & 0.05 & 0.049 & 0.057 \\ 
 & 192 & 0.103 & 0.072 & 0.065 & 0.105 & 0.095 & \textbf{0.047} & \uline{0.055} & 0.243 \\ 
\hline 
\multirow{5}*{\rotatebox{90}{Electricity}} & 24 & \uline{0.409} & 0.421 & 0.433 & 0.463 & 0.429 & 0.42 & \uline{0.409} & \textbf{0.384} \\ 
 & 48 & \uline{0.28} & 0.308 & 0.304 & 0.338 & 0.313 & 0.289 & \textbf{0.278} & 0.351 \\ 
 & 60 & \textbf{0.26} & 0.287 & 0.279 & 0.346 & 0.301 & 0.279 & \uline{0.273} & 0.363 \\ 
 & 96 & \uline{0.231} & 0.243 & 0.238 & 0.352 & 0.294 & \uline{0.231} & \textbf{0.227} & 0.317 \\ 
 & 192 & \uline{0.495} & \uline{0.495} & \textbf{0.491} & 0.616 & 0.548 & 0.503 & 0.507 & 0.551 \\ 
\hline 
\multirow{5}*{\rotatebox{90}{Traffic}} & 24 & \textbf{0.231} & \uline{0.232} & 0.239 & 0.324 & 0.3 & 0.288 & 0.301 & 0.237 \\ 
 & 48 & 0.193 & \textbf{0.183} & \uline{0.187} & 0.244 & 0.232 & 0.213 & 0.226 & 0.192 \\ 
 & 60 & 0.211 & \textbf{0.191} & \uline{0.196} & 0.258 & 0.328 & 0.332 & 0.326 & 0.198 \\ 
 & 96 & 0.21 & \uline{0.199} & 0.215 & 0.256 & 0.219 & 0.217 & \textbf{0.196} & 0.213 \\ 
 & 192 & 0.387 & 0.393 & 0.382 & 0.373 & 0.419 & \uline{0.351} & \textbf{0.297} & 0.374 \\ 
\hline 
 \end{tabular}
   }
    \captionof{table}{Zero-shot results of TTM over Chronos (2024) and Lag-llama (2024) over the last test-window. Since Chronos and Lag-llama recommend/report results with shorter forecast lengths, we use different $FLs \in \{24,48,60,96,192\}$ in this experiment.}
    \label{tab:n_zs_2_full}
  \end{minipage}%
\end{figure*}


\begin{figure*}[h]
  \begin{minipage}{0.95\textwidth}
    \centering
    \setlength{\tabcolsep}{2pt}
    \scalebox{.75}{

    \begin{tabular}{|c|c|c|c|c|c|c|c|c|c|c|c|c|c|} \hline 
    \makecell{Data} & & \multicolumn{5}{|c|}{\makecell{\textbf{Pre-trained Models}}} & \multicolumn{6}{|c|}{\makecell{\textbf{Other popular architectures}}} \\ \hline 
 & \makecell{FL} & \makecell{\ttms~} & \makecell{\ttmm~} & \makecell{\ttml~} & \makecell{GPT4TS} & \makecell{Time-LLM} & \makecell{PatchTST} & \makecell{TSMixer} & \makecell{TimeMixer} & \makecell{iTransformer} & \makecell{TimesNet} & \makecell{Dlinear}\\ \hline 
\multirow{4}*{\rotatebox{90}{ETTH1}} & 96 & 0.364 & \uline{0.363} & \textbf{0.359} & 0.543 & 0.483 & 0.557 & 0.554 & 0.899 & 0.674 & 0.892 & 0.547 \\ 
 & 192 & \textbf{0.387} & \uline{0.393} & 0.394 & 0.748 & 0.629 & 0.711 & 0.673 & 0.942 & 0.757 & 0.94 & 0.72 \\ 
 & 336 & \uline{0.399} & \textbf{0.398} & 0.406 & 0.754 & 0.768 & 0.816 & 0.678 & 1.423 & 0.838 & 0.945 & 0.984 \\ 
 & 720 & - & - & - & - & - & - & - & - & - & - & - \\ 
\hline 
\multirow{4}*{\rotatebox{90}{ETTH2}} & 96 & 0.277 & \uline{0.271} & \textbf{0.267} & 0.376 & 0.336 & 0.401 & 0.348 & 0.356 & 0.382 & 0.409 & 0.442 \\ 
 & 192 & 0.334 & \uline{0.325} & \textbf{0.321} & 0.418 & 0.406 & 0.452 & 0.419 & 0.549 & 0.445 & 0.483 & 0.617 \\ 
 & 336 & 0.361 & \uline{0.357} & \textbf{0.354} & 0.408 & 0.405 & 0.464 & 0.389 & 0.619 & 0.483 & 0.499 & 1.424 \\ 
 & 720 & - & - & - & - & - & - & - & - & - & - & - \\ 
\hline 
\multirow{4}*{\rotatebox{90}{ETTM1}} & 96 & \textbf{0.313} & 0.326 & 0.317 & 0.386 & \uline{0.316} & 0.399 & 0.361 & 0.515 & 0.437 & 0.606 & 0.332 \\ 
 & 192 & \uline{0.357} & 0.371 & \textbf{0.355} & 0.44 & 0.45 & 0.441 & 0.411 & 0.535 & 0.49 & 0.681 & 0.358 \\ 
 & 336 & \uline{0.395} & 0.396 & \textbf{0.374} & 0.485 & 0.45 & 0.499 & 0.467 & 0.621 & 0.563 & 0.786 & 0.402 \\ 
 & 720 & 0.437 & \uline{0.418} & \textbf{0.397} & 0.577 & 0.483 & 0.767 & 0.677 & 0.64 & 0.78 & 0.796 & 0.511 \\ 
\hline 
\multirow{4}*{\rotatebox{90}{ETTM2}} & 96 & \uline{0.171} & 0.178 & \textbf{0.17} & 0.199 & 0.174 & 0.206 & 0.2 & 0.231 & 0.217 & 0.22 & 0.236 \\ 
 & 192 & 0.23 & 0.237 & \uline{0.222} & 0.256 & \textbf{0.215} & 0.264 & 0.265 & 0.313 & 0.266 & 0.311 & 0.306 \\ 
 & 336 & 0.293 & 0.284 & \uline{0.274} & 0.318 & \textbf{0.273} & 0.334 & 0.314 & 0.356 & 0.322 & 0.338 & 0.38 \\ 
 & 720 & 0.393 & \uline{0.373} & \textbf{0.345} & 0.46 & 0.433 & 0.454 & 0.41 & 0.46 & 0.43 & 0.509 & 0.674 \\ 
\hline 
\multirow{4}*{\rotatebox{90}{Weather}} & 96 & \textbf{0.154} & 0.162 & \uline{0.155} & 0.175 & 0.172 & 0.171 & 0.188 & 0.187 & 0.211 & 0.207 & 0.184 \\ 
 & 192 & \uline{0.203} & 0.215 & \textbf{0.201} & 0.227 & 0.224 & 0.23 & 0.234 & 0.324 & 0.269 & 0.272 & 0.228 \\ 
 & 336 & \uline{0.252} & 0.262 & \textbf{0.244} & 0.286 & 0.282 & 0.294 & 0.287 & 0.307 & 0.316 & 0.313 & 0.279 \\ 
 & 720 & 0.327 & \uline{0.319} & \textbf{0.316} & 0.366 & 0.366 & 0.384 & 0.365 & 0.451 & 0.393 & 0.4 & 0.364 \\ 
\hline 
\multirow{4}*{\rotatebox{90}{Electricity}} & 96 & 0.146 & 0.15 & \textbf{0.141} & \uline{0.143} & 0.147 & 0.145 & 0.147 & 0.155 & 0.157 & 0.315 & 0.15 \\ 
 & 192 & 0.164 & 0.171 & 0.16 & \uline{0.159} & \textbf{0.158} & 0.163 & 0.172 & 0.219 & 0.181 & 0.318 & 0.163 \\ 
 & 336 & 0.185 & 0.202 & 0.179 & 0.179 & \uline{0.178} & \textbf{0.175} & 0.19 & 0.273 & 0.219 & 0.34 & \textbf{0.175} \\ 
 & 720 & 0.236 & 0.303 & 0.24 & 0.233 & \uline{0.224} & \textbf{0.219} & 0.28 & 0.309 & 0.249 & 0.635 & \textbf{0.219} \\ 
\hline 
\multirow{4}*{\rotatebox{90}{Traffic}} & 96 & 0.411 & 0.411 & 0.469 & 0.419 & 0.414 & \textbf{0.404} & \uline{0.408} & 0.463 & 0.43 & 0.854 & 0.427 \\ 
 & 192 & 0.42 & 0.44 & 0.488 & 0.434 & \uline{0.419} & \textbf{0.412} & 0.421 & 0.548 & 0.445 & 0.894 & 0.447 \\ 
 & 336 & 0.468 & 0.46 & 0.512 & 0.449 & \textbf{0.437} & \uline{0.439} & 0.477 & 0.498 & 0.481 & 0.853 & 0.478 \\ 
 & 720 & - & - & - & - & - & - & - & - & - & - & - \\ 
\hline 
 \end{tabular}
    
    }
    \captionof{table}{TTM Few-shot 5\% MSE reported across all the standard FLs considered. TTM, Pre-trained baselines and other model architectures are trained with 5\% train data. }
    \label{tab:n_t3_fs5_full}
  \end{minipage}%
\end{figure*}


\begin{figure*}[h]
  \begin{minipage}{0.95\textwidth}
    \centering
    \setlength{\tabcolsep}{2pt}
    \scalebox{.63}{

   \begin{tabular}{|c|c|c|c|c|c|c|c|c|c|c|c|c|c|c|c|} \hline 

\makecell{Data} & & \multicolumn{6}{|c|}{\makecell{\textbf{Head Probing of Pre-Trained Models}}} & \multicolumn{7}{|c|}{\makecell{\textbf{Full-shot end-2-end Training with 512 context length}}} \\ \hline 

 & \makecell{FL} & \makecell{\ttms~} & \makecell{\ttmm~} & \makecell{\ttml~} & \makecell{Moment} & \makecell{GPT4TS} & \makecell{Time-LLM} & \makecell{TimeMixer} & \makecell{PatchTST} & \makecell{TimesNet} & \makecell{TSMixer} & \makecell{FEDFormer} & \makecell{Autoformer} & \makecell{Informer}\\ \hline 
\multirow{2}*{\rotatebox{0}{ETTH1}} & 96 & \textbf{0.36} & 0.362 & 0.363 & 0.387 & 0.376 & 0.408 & \uline{0.361} & 0.37 & 0.384 & 0.368 & 0.376 & 0.449 & 0.865 \\ 
 & 720 & \textbf{0.436} & 0.449 & \uline{0.442} & 0.454 & 0.477 & 0.523 & 0.445 & 0.447 & 0.521 & 0.444 & 0.506 & 0.514 & 1.181 \\ 
\hline 
\multirow{2}*{\rotatebox{0}{ETTH2}} & 96 & \uline{0.269} & 0.273 & \textbf{0.262} & 0.288 & 0.285 & 0.285 & 0.271 & 0.274 & 0.34 & 0.276 & 0.346 & 0.358 & 3.755 \\ 
 & 720 & 0.39 & 0.402 & 0.392 & 0.403 & 0.406 & 0.399 & \textbf{0.342} & \uline{0.379} & 0.462 & 0.395 & 0.463 & 0.515 & 3.647 \\ 
\hline 
\multirow{2}*{\rotatebox{0}{ETTM1}} & 96 & 0.291 & 0.293 & \textbf{0.283} & 0.293 & 0.292 & 0.384 & 0.291 & \uline{0.29} & 0.338 & 0.291 & 0.379 & 0.505 & 0.672 \\ 
 & 720 & 0.419 & 0.408 & \textbf{0.393} & \uline{0.405} & 0.417 & 0.437 & 0.415 & 0.416 & 0.478 & 0.416 & 0.543 & 0.671 & 1.166 \\ 
\hline 
\multirow{2}*{\rotatebox{0}{ETTM2}} & 96 & \uline{0.164} & \textbf{0.158} & \textbf{0.158} & 0.17 & 0.173 & 0.181 & \uline{0.164} & 0.165 & 0.187 & \uline{0.164} & 0.203 & 0.255 & 0.365 \\ 
 & 720 & \uline{0.35} & \textbf{0.347} & 0.369 & 0.363 & 0.378 & 0.366 & 0.359 & 0.362 & 0.408 & 0.358 & 0.421 & 0.433 & 3.379 \\ 
\hline 
\multirow{2}*{\rotatebox{0}{Weather}} & 96 & \textbf{0.146} & 0.154 & \uline{0.149} & 0.154 & 0.162 & - & 0.147 & \uline{0.149} & 0.172 & \textbf{0.146} & 0.217 & 0.266 & 0.3 \\ 
 & 720 & 0.323 & 0.324 & 0.318 & 0.315 & 0.326 & - & \textbf{0.31} & \uline{0.314} & 0.365 & 0.316 & 0.403 & 0.419 & 1.059 \\ 
\hline 
\multirow{2}*{\rotatebox{0}{Electricity}} & 96 & \uline{0.129} & \uline{0.129} & \textbf{0.128} & 0.138 & 0.139 & - & \uline{0.129} & \uline{0.129} & 0.168 & \uline{0.129} & 0.193 & 0.201 & 0.274 \\ 
 & 720 & 0.2 & 0.193 & \uline{0.191} & 0.211 & 0.206 & - & 0.194 & 0.197 & 0.22 & \textbf{0.186} & 0.246 & 0.254 & 0.373 \\ 
\hline 
\multirow{2}*{\rotatebox{0}{Traffic}} & 96 & 0.368 & 0.372 & \textbf{0.352} & 0.391 & 0.388 & - & 0.36 & 0.36 & 0.593 & \uline{0.356} & 0.587 & 0.613 & 0.719 \\ 
 & 720 & \textbf{0.431} & 0.425 & 0.419 & 0.45 & 0.45 & - & 0.43 & 0.432 & 0.64 & 0.424 & 0.626 & 0.66 & 0.864 \\ 
\hline 

 \end{tabular}
 
   }
    \captionof{table}{Head Probing (HP) involves finetuning the pre-trained model heads on full data with backbone weights frozen. Head probing results of TTM are compared with the Head probing results of other Pre-trained models and also with the Full-shot end-to-end training of popular TS architectures. TTM's Head probing results consistently outperform other HP benchmarks and also very competitive as compared to the full end-to-end training of popular TS architectures. MSE across FLs (96,720) are reported from \cite{moment}. Time-LLM results for large datasets are not reported in \cite{moment} due to computational issues. It is important to note that end-to-end training updates the backbone weights, whereas head probing does not.}
    \label{tab:n_hp_full}
  \end{minipage}%
\end{figure*}

\begin{figure*}[h]
  \begin{minipage}{0.95\textwidth}
    \centering
    \setlength{\tabcolsep}{2pt}
    \scalebox{.7}{

    \begin{tabular}{|c|c|c|c|c|c|c|c|c|c|c|c|c|c|c|c|} \hline 

    \makecell{Data} & & \multicolumn{4}{|c|}{\makecell{\textbf{TTM Head Probing}}} & \multicolumn{9}{|c|}{\makecell{\textbf{Full-shot end-2-end Training with 512 context length}}} \\ \hline 
    
 & \makecell{FL} & \makecell{\ttmq~} & \makecell{\ttms~} & \makecell{\ttmm~} & \makecell{\ttml~} & \makecell{TimeMixer} & \makecell{TSMixer} & \makecell{PatchTST} & \makecell{TimesNet} & \makecell{CrossFormer} & \makecell{Dlinear} & \makecell{FEDFormer} & \makecell{Autoformer} & \makecell{Informer}\\ \hline 
\multirow{4}*{\rotatebox{90}{ETTH1}} & 96 & 0.373 & \textbf{0.36} & 0.362 & 0.363 & \uline{0.361} & 0.368 & 0.37 & 0.384 & 0.418 & 0.375 & 0.376 & 0.449 & 0.865 \\ 
 & 720 & \textbf{0.424} & \uline{0.436} & 0.449 & 0.442 & 0.445 & 0.444 & 0.447 & 0.521 & 0.733 & 0.472 & 0.506 & 0.514 & 1.181 \\ 
 & 192 & 0.398 & \textbf{0.392} & \uline{0.394} & \textbf{0.392} & 0.409 & 0.399 & 0.413 & 0.436 & 0.539 & 0.405 & 0.42 & 0.5 & 1.008 \\ 
 & 336 & \textbf{0.397} & \uline{0.401} & 0.403 & 0.413 & 0.43 & 0.421 & 0.422 & 0.638 & 0.709 & 0.439 & 0.459 & 0.521 & 1.107 \\ 
\hline 
\multirow{4}*{\rotatebox{90}{ETTH2}} & 96 & 0.283 & \uline{0.269} & 0.273 & \textbf{0.262} & 0.271 & 0.276 & 0.274 & 0.34 & 0.425 & 0.289 & 0.346 & 0.358 & 3.755 \\ 
 & 720 & 0.417 & 0.39 & 0.402 & 0.392 & \textbf{0.342} & 0.395 & \uline{0.379} & 0.462 & 0.775 & 0.605 & 0.463 & 0.515 & 3.647 \\ 
 & 192 & 0.328 & 0.336 & 0.325 & 0.324 & 0.317 & 0.33 & \uline{0.314} & \textbf{0.231} & 0.473 & 0.383 & 0.429 & 0.456 & 5.602 \\ 
 & 336 & 0.361 & 0.359 & 0.356 & 0.351 & \uline{0.332} & 0.357 & \textbf{0.329} & 0.452 & 0.581 & 0.448 & 0.496 & 0.482 & 4.721 \\ 
\hline 
\multirow{4}*{\rotatebox{90}{ETTM1}} & 96 & \uline{0.286} & 0.291 & 0.293 & \textbf{0.283} & 0.291 & 0.291 & 0.293 & 0.338 & 0.361 & 0.299 & 0.379 & 0.505 & 0.672 \\ 
 & 720 & 0.417 & 0.419 & \uline{0.408} & \textbf{0.393} & 0.415 & 0.416 & 0.416 & 0.478 & 0.703 & 0.425 & 0.543 & 0.671 & 1.166 \\ 
 & 192 & 0.333 & \textbf{0.325} & 0.335 & 0.332 & \uline{0.327} & 0.333 & 0.333 & 0.374 & 0.387 & 0.335 & 0.426 & 0.553 & 0.795 \\ 
 & 336 & 0.364 & 0.363 & 0.364 & \textbf{0.353} & \uline{0.36} & 0.365 & 0.369 & 0.41 & 0.605 & 0.369 & 0.445 & 0.621 & 1.212 \\ 
\hline 
\multirow{4}*{\rotatebox{90}{ETTM2}} & 96 & 0.165 & \uline{0.164} & \textbf{0.158} & \textbf{0.158} & \uline{0.164} & \uline{0.164} & 0.166 & 0.187 & 0.275 & 0.167 & 0.203 & 0.255 & 0.365 \\ 
 & 720 & 0.358 & \uline{0.35} & \textbf{0.347} & 0.369 & 0.359 & 0.358 & 0.362 & 0.408 & 1.208 & 0.397 & 0.421 & 0.433 & 3.379 \\ 
 & 192 & 0.22 & 0.219 & \uline{0.215} & \textbf{0.213} & 0.223 & 0.219 & 0.223 & 0.249 & 0.345 & 0.224 & 0.269 & 0.281 & 0.533 \\ 
 & 336 & \uline{0.269} & 0.277 & \textbf{0.26} & \uline{0.269} & 0.279 & 0.273 & 0.274 & 0.321 & 0.657 & 0.281 & 0.325 & 0.339 & 1.363 \\ 
\hline 
\multirow{4}*{\rotatebox{90}{Weather}} & 96 & \textbf{0.144} & \uline{0.146} & 0.154 & 0.149 & 0.147 & \uline{0.146} & 0.149 & 0.172 & 0.232 & 0.176 & 0.217 & 0.266 & 0.3 \\ 
 & 720 & 0.317 & 0.323 & 0.324 & 0.318 & \textbf{0.31} & 0.316 & \uline{0.314} & 0.365 & 0.526 & 0.323 & 0.403 & 0.419 & 1.059 \\ 
 & 192 & \uline{0.19} & \uline{0.19} & 0.207 & 0.192 & \textbf{0.189} & 0.191 & 0.194 & 0.219 & 0.371 & 0.22 & 0.276 & 0.307 & 0.598 \\ 
 & 336 & 0.247 & 0.242 & 0.25 & \textbf{0.24} & \uline{0.241} & 0.243 & 0.306 & 0.246 & 0.495 & 0.265 & 0.339 & 0.359 & 0.578 \\ 
\hline 
\multirow{4}*{\rotatebox{90}{Electricity}} & 96 & 0.13 & \uline{0.129} & \uline{0.129} & \textbf{0.128} & \uline{0.129} & \uline{0.129} & \uline{0.129} & 0.168 & 0.15 & 0.14 & 0.193 & 0.201 & 0.274 \\ 
 & 720 & 0.202 & 0.2 & 0.193 & \uline{0.191} & 0.194 & \textbf{0.186} & 0.197 & 0.22 & 0.251 & 0.203 & 0.246 & 0.254 & 0.373 \\ 
 & 192 & 0.149 & 0.149 & 0.148 & \uline{0.144} & \textbf{0.14} & 0.146 & 0.147 & 0.184 & 0.161 & 0.153 & 0.201 & 0.222 & 0.296 \\ 
 & 336 & 0.164 & 0.163 & \uline{0.161} & 0.162 & \uline{0.161} & \textbf{0.158} & 0.163 & 0.198 & 0.182 & 0.169 & 0.214 & 0.213 & 0.3 \\ 
\hline 
\multirow{4}*{\rotatebox{90}{Traffic}} & 96 & 0.367 & 0.368 & 0.372 & \textbf{0.352} & 0.36 & \uline{0.356} & 0.36 & 0.593 & 0.514 & 0.41 & 0.587 & 0.613 & 0.719 \\ 
 & 720 & 0.432 & 0.431 & 0.425 & \textbf{0.419} & 0.43 & \uline{0.424} & 0.432 & 0.64 & 0.573 & 0.466 & 0.626 & 0.66 & 0.864 \\ 
 & 192 & 0.387 & 0.403 & \uline{0.365} & \textbf{0.359} & 0.375 & 0.377 & 0.379 & 0.617 & 0.549 & 0.423 & 0.604 & 0.616 & 0.696 \\ 
 & 336 & 0.414 & 0.395 & \uline{0.379} & \textbf{0.375} & 0.385 & 0.385 & 0.392 & 0.629 & 0.53 & 0.436 & 0.621 & 0.622 & 0.777 \\ 
\hline 
 \end{tabular}
   
   }
    \captionof{table}{\textbf{Full view: TTM Head probing Vs Full-shot End-To-End Training} of popular time-series architectures reported for all FLs. Head Probing (HP) involves finetuning the pre-trained model heads on full data with backbone weights frozen. TTM head probing results are either superior or highly competitive with other popular state-of-the-art methods that are trained end-to-end on the target data. It is important to note that end-to-end training updates the backbone weights, whereas head probing does not.}
    \label{tab:n_hp_vs_fulle2e_all_fls}
  \end{minipage}%
\end{figure*}

\begin{figure*}[h]
  \begin{minipage}{0.95\textwidth}
    \centering
    \setlength{\tabcolsep}{2pt}
    \scalebox{.7}{

    \begin{tabular}{|c|c|c|c|c|c|c|c|c|c|c|c|c|c|} \hline 

    \makecell{Data} & \multicolumn{4}{|c|}{\makecell{\textbf{TTM Head Probing}}} & \multicolumn{9}{|c|}{\makecell{\textbf{Full-shot end-2-end Training with 512 context length}}} \\ \hline 
    
  \makecell{Data} & \makecell{\ttmq~} & \makecell{\ttms~} & \makecell{\ttmm~} & \makecell{\ttml~} & \makecell{TimeMixer} & \makecell{TSMixer} & \makecell{PatchTST} & \makecell{TimesNet} & \makecell{CrossFormer} & \makecell{Dlinear} & \makecell{FEDFormer} & \makecell{Autoformer} & \makecell{Informer}\\ \hline 

ETTH1 & \uline{0.398} & \textbf{0.397} & 0.402 & 0.402 & 0.411 & 0.408 & 0.413 & 0.495 & 0.6 & 0.423 & 0.44 & 0.496 & 1.04 \\ 
ETTH2 & 0.347 & 0.338 & 0.339 & 0.332 & \textbf{0.316} & 0.34 & \uline{0.324} & 0.371 & 0.564 & 0.431 & 0.434 & 0.453 & 4.431 \\ 
ETTM1 & 0.35 & 0.35 & 0.35 & \textbf{0.34} & \uline{0.348} & 0.351 & 0.353 & 0.4 & 0.514 & 0.357 & 0.448 & 0.588 & 0.961 \\ 
ETTM2 & 0.253 & \uline{0.252} & \textbf{0.245} & \uline{0.252} & 0.256 & 0.254 & 0.256 & 0.291 & 0.621 & 0.267 & 0.304 & 0.327 & 1.41 \\ 
Weather & \uline{0.224} & 0.225 & 0.234 & 0.225 & \textbf{0.222} & \uline{0.224} & 0.241 & 0.25 & 0.406 & 0.246 & 0.309 & 0.338 & 0.634 \\ 
Electricity & 0.161 & 0.16 & 0.158 & \uline{0.156} & \uline{0.156} & \textbf{0.155} & 0.159 & 0.192 & 0.186 & 0.166 & 0.214 & 0.222 & 0.311 \\ 
Traffic & 0.4 & 0.399 & \uline{0.385} & \textbf{0.376} & 0.388 & 0.386 & 0.391 & 0.62 & 0.542 & 0.434 & 0.609 & 0.628 & 0.764 \\ 
\hline 
\end{tabular}

   }
    \captionof{table}{\textbf{Average view: TTM Head probing Vs Full-shot End-To-End Training} of popular time-series architectures averaged across FLs {96,192,336,720}. Head Probing (HP) involves finetuning the pre-trained model heads on full data with backbone weights frozen. TTM head probing results are either superior or highly competitive with other popular state-of-the-art methods that are trained end-to-end on the target data. It is important to note that end-to-end training updates the backbone weights, whereas head probing does not.}
    \label{tab:n_hp_vs_fulle2e_avg_fls}
  \end{minipage}%
\end{figure*}

\section{Sample Zero-shot Visualizations}
\begin{figure*}
\centering 
\includegraphics[width=0.95\textwidth]{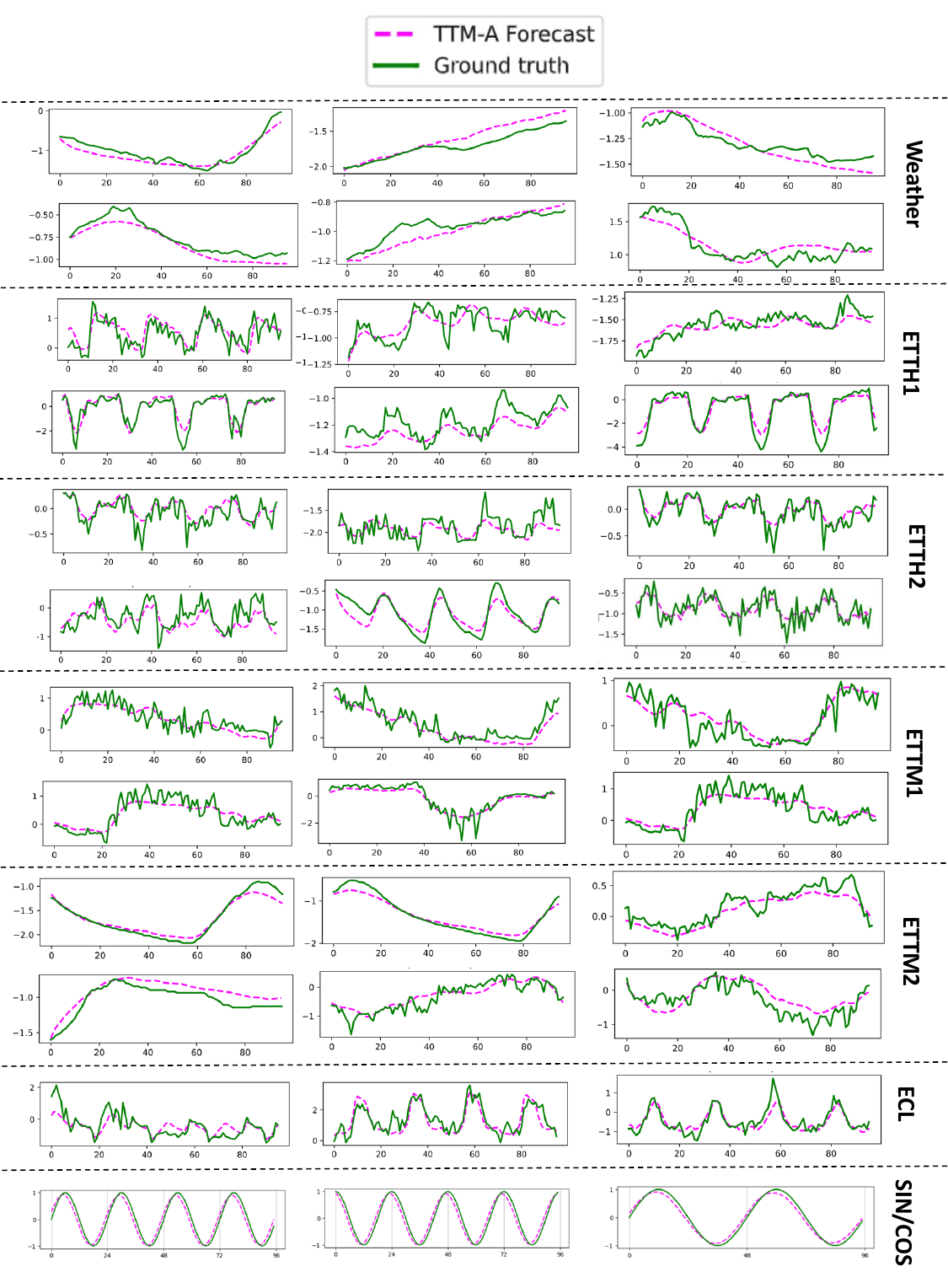}
\caption{Sample TTM Zero-shot Forecasts across datasets}
\label{fig:sample_viz}
\end{figure*}

Figure.~\ref{fig:sample_viz} visualizes the zero-shot forecasts of TTM across different datasets illustrating the power of TTM to capture complex trends and seasonal patterns.

\section{Full Results Tables}
\label{appendix:full_tables}
Here, we present the complete versions of various tables in the main paper. These full versions essentially include the test results for multiple forecast lengths ($fl$) across all datasets. Occasionally, these results are averaged across forecast lengths to conserve space in the main paper.

\subsection{Full table for all TTM variants}
Table~\ref{tab:appendix_zs_master} and Table~\ref{tab:appendix_fs5_master} captures the fine-grained results of all TTM variants (i.e. \ttmq, \ttms, \ttmm and \ttml) on the $D1$ data benchmark set.

\subsection{Full table for zero-shot experiment}
\label{appendix:zs}
Table~\ref{tab:n_zs_moirai_full} show the sliding window zero-shot results for all forecast lengths across all D1 datasets, and compares TTM variants with Moirai variants and TimesFM.
Table~\ref{tab:n_zs_2_full} depicts the last-window zero-shot results for all forecast lengths across all of D1 datasets, and compares TTM with Chronos and Lag-Llama.

\subsection{Other zero-shot comparisons}
\label{appendix:llmtime_unitime}
LLMTime~\cite{llmtime} reported the test performance only on the last windows of the test datasets (instead of sliding windows) for horizons 96 and 192 due to computational reasons. 
We recreate the same experimental setup for TTM, and depict the comparative results in Table~\ref{tab:n_zs_llmtime_full}.
We observe 26-36\% improvement across all three variants of TTM with tremendous (70,000 to 14,000) reduction in model sizes.
Another zero-shot comparison with the UniTime~\cite{unitime} model is shown in Table~\ref{tab:unitime}. In this comparison,
TTM outperforms UniTime by 29-31\%. 

\subsection{TTM Zero-shot \textit{vs.} SOTA Full-shot (short context setting)}
\label{appendix:zs_fullshot_short}
In Table~\ref{tab:n_zs_moirai_fullshot_avg} we compare the zero-shot results of TTM variants with full-shot end-2-end training of popular TS architectures like iTransformer, PatchTST \etc.
The full-shot SOTA algorithms were trained in short-context length setting ($sl=96$) on the train split of each target dataset, and these results are obtained from the Moirai paper~\cite{moirai} where the authors draw similar comparison.
TTM was tested in zero-shot setting without any training on the target datasets.
We also provide the zero-shot results of Moirai variants and TimesFM from Table~\ref{tab:n_zs_moirai_avg} for reference purpose.
We can see that the zero-shot performance of all variants of TTM outperforms the full-shot performance of SOTA models, even though the latter are trained on the target datasets. This underscores the strength of the pre-trained TTM model.

\subsection{Full table for 5\% few-shot experiment}
\label{appendix:fs5}
Table~\ref{tab:n_t3_fs5_full} shows the 5\% few-shot results for all forecast lengths across all D1 datasets.

\subsection{TTM \textit{vs.} Cross-transfer models}
\label{appendix:cross_transfer}
Table~\ref{tab:simmtm_2} draws a comparative analysis of \ttmq~with SimMTM, Ti-MAE, TST, LaST, TF-C, CoST, and TS2Vec models in different few-shot settings (10\% to 100\% availability of training data) on ETTH1 dataset.
The baseline models are trained on ETTH2 data, and tested on ETTH1 data, thus demonstrating their transferability across datasets having similar characteristics.
The baseline numbers are taken from~\cite{dong2023simmtm}.
\ttmq~outperform all of them (including the recent SOTA SimMTM) by a significant margin.
This highlights the usefulness of the pre-trained TTM weights and their ability to adapt to a target domain with few-shot fine-tuning.

\subsection{Full table for full-shot head-probing experiment}
\label{appendix:hp_full_shot_512}
Head Probing (HP) involves finetuning the pre-trained model heads on full data with backbone weights frozen. Table~\ref{tab:n_hp_full} compares the full-shot head-probing results of TTM with Moment, Time-LLM and GPT4TS. Table~\ref{tab:n_hp_vs_fulle2e_all_fls} and Table~\ref{tab:n_hp_vs_fulle2e_avg_fls} compares the TTM Head probing results with the Full-shot End-To-End Training results of popular time-series architectures.  It is important to note that end-to-end training updates the backbone weights, whereas head probing does not. TTM head probing results are either superior or highly competitive with other popular state-of-the-art methods that are trained end-to-end on the target data.

\subsection{Full table: Impact of Adaptive Patching (AP)}
\label{appendix:AP}
Table~\ref{tab:abl_ap} shows the full table for studying the impact of adaptive patching across different amounts of pre-trained data settings.
In both the settings, AP helps TTM to produce more accurate forecasts.
However, the impact of AP is greater in the setting with a lesser amount of pre-training data, where there is more need to model at multiple granularities to compensate for the data size.

\subsection{Full table: Impact of Resolution Prefix Tuning (RPT)}
\label{appendix:RPT}
Table~\ref{tab:abl_fpt} presents a comprehensive analysis of the impact of RPT on TTM. RPT generally enhances forecast performance, particularly when the pretraining (PT) data is abundant and diverse. In this scenario, incorporating a learnable resolution prefix token significantly benefits the models by allowing them to decouple the weights across resolutions effectively. Conversely, in setups with limited PT data where diversity challenges are minimal, RPT has a reduced impact.
Additionally, we can see that RPT helps in scenarios when the context length is short. Table~\ref{tab:abl_rpt_2} shows the impact of RPT in shorter context length setting ($sl=96$).
We report the zero-shot results for $fl=24$ in the table. 
 In these scenarios, automatically detecting the resolution becomes a challenge for the model. Hence, by explicitly fusing the resolution information as a prefix, we can enhance the model’s ability to learn effectively across resolutions.

\begin{figure*}[h]
  \begin{minipage}{0.95\textwidth}
    \centering
    \scalebox{.75}{
\begin{tabular}{|c|c|c|c|c|c|c|} \hline 
 & \makecell{FL} & \makecell{\ttms~} & \makecell{\ttmm~} & \makecell{\ttml~} & \makecell{LLMTime}\\ \hline 
\multirow{2}*{\rotatebox{0}{ETTM2}} & 96 & 0.137 & \uline{0.108} & \textbf{0.101} & 0.167 \\ 
 & 192 & 0.175 & \uline{0.165} & \textbf{0.146} & 0.198 \\ 
\hline 
\multirow{2}*{\rotatebox{0}{Weather}} & 96 & \uline{0.034} & \textbf{0.025} & 0.053 & 0.107 \\ 
 & 192 & 0.103 & 0.072 & \uline{0.065} & \textbf{0.062} \\ 
\hline 
\multirow{2}*{\rotatebox{0}{Electricity}} & 96 & \textbf{0.231} & 0.243 & \uline{0.238} & 0.609 \\ 
 & 192 & \uline{0.495} & \uline{0.495} & \textbf{0.491} & 0.96 \\ 
\hline 
\multirow{2}*{\rotatebox{0}{Traffic}} & 96 & \uline{0.21} & \textbf{0.199} & 0.215 & 0.34 \\ 
 & 192 & \uline{0.387} & 0.393 & \textbf{0.382} & 0.526 \\ 
\hline 
\multicolumn{2}{|c|}{\makecell{\textbf{Model Size}}} & \textbf{1M} & \textbf{4M} & \textbf{5M} & \textbf{70B}\\ \hline 
\multicolumn{5}{|c|}{\makecell{\textbf{\ttms~} \textit{f-imp(\%) s-imp(X)}}} & \textbf{26\% $\uparrow$ 70000X $\uparrow$} \\ 
\multicolumn{5}{|c|}{\makecell{\textbf{\ttmm~} \textit{f-imp(\%) s-imp(X)}}} & \textbf{36\% $\uparrow$ 17500X $\uparrow$} \\ 
\multicolumn{5}{|c|}{\makecell{\textbf{\ttml~} \textit{f-imp(\%) s-imp(X)}}} & \textbf{36\% $\uparrow$ 14000X $\uparrow$} \\ 
\hline 
 \end{tabular}
   }
    \captionof{table}{LLM-Time Vs TTM: Zeroshot MSE reported on last test window set. Results reported in LLMTime~\cite{llmtime} are used for this comparison.}
    \label{tab:n_zs_llmtime_full}
  \end{minipage}%
\end{figure*}


\begin{figure*}[h]
  \begin{minipage}{0.95\textwidth}

  \centering
    \scalebox{0.75}{
    \begin{tabular}{|c|c|c|c|c|c|}
    \hline
    & \makecell{FL} & \makecell{\ttms~} & \makecell{\ttmm~} & \makecell{\ttml~} & \makecell{Unitime}\\ \hline 
    \multirow{4}*{\rotatebox{90}{ETTH2}} & 96 & 0.277 & \uline{0.271} & \textbf{0.264} & 0.306 \\ 
     & 192 & 0.334 & \uline{0.324} & \textbf{0.321} & 0.389 \\ 
     & 336 & 0.362 & \uline{0.357} & \textbf{0.351} & 0.424 \\ 
     & 720 & 0.408 & \textbf{0.388} & \uline{0.395} & 0.434 \\ 
    \hline 
    \multirow{4}*{\rotatebox{90}{Weather}} & 96 & \textbf{0.158} & 0.166 & \uline{0.159} & 0.21 \\ 
     & 192 & \uline{0.206} & 0.214 & \textbf{0.203} & 0.264 \\ 
     & 336 & 0.256 & \uline{0.254} & \textbf{0.247} & 0.326 \\ 
     & 720 & 0.328 & \uline{0.319} & \textbf{0.314} & 0.402 \\ 
    \hline 
    \multirow{4}*{\rotatebox{90}{Electricity}} & 96 & 0.166 & \uline{0.157} & \textbf{0.152} & 0.409 \\ 
     & 192 & 0.191 & \textbf{0.174} & \uline{0.179} & 0.41 \\ 
     & 336 & 0.207 & \uline{0.195} & \textbf{0.193} & 0.439 \\ 
     & 720 & 0.255 & \uline{0.25} & \textbf{0.243} & 0.487 \\ 
    \hline 
    \multicolumn{2}{|c|}{\makecell{\textbf{\ttms~f-imp(\%)}}} & \multicolumn{4}{|c|}{29\%} \\ 
    \multicolumn{2}{|c|}{\makecell{\textbf{\ttmm~f-imp(\%)}}} & \multicolumn{4}{|c|}{30\%} \\
    \multicolumn{2}{|c|}{\makecell{\textbf{\ttml~f-imp(\%)}}} & \multicolumn{4}{|c|}{31\%} \\ \hline 

    \end{tabular}}
    \captionsetup{justification=centering,singlelinecheck=false}
    \captionof{table}{TTM vs UniTime MSE Improvement (\textit{f-imp}) in zero-shot setting using full sliding-window test set. Results reported in UniTime~\cite{unitime} are used for this comparison.}
    \label{tab:unitime}
    
  \end{minipage}
  
\end{figure*}

\begin{figure*}[h]

\begin{minipage}{0.8\textwidth}

   \centering
    \setlength{\tabcolsep}{1.5pt}
    \scalebox{0.7}{
    \begin{tabular}{|c|c|c|c|c|c|c|}
    \hline
    \makecell{} & \makecell{10\%} & \makecell{25\%} & \makecell{50\%} & \makecell{75\%} & \makecell{100\%} & \makecell{\textbf{IMP}}\\ \hline 
    \textbf{\ttmq} & \textbf{0.422} & \textbf{0.421} & \textbf{0.413} & \textbf{0.402} & \textbf{0.398} & - \\ 
    \makecell{SimMTM\\(NeurIPS 23)} & 0.591 & 0.535 & 0.491 & 0.466 & 0.415 & \textbf{17\%} \\ 
    Ti-MAE & 0.660 & 0.594 & 0.55 & 0.475 & 0.466 & \textbf{24\%} \\ 
    TST & 0.783 & 0.641 & 0.525 & 0.516 & 0.469 & \textbf{28\%} \\ 
    LaST & 0.645 & 0.610 & 0.540 & 0.479 & 0.443 & \textbf{23\%} \\ 
    TF-C & 0.799 & 0.736 & 0.731 & 0.697 & 0.635 & \textbf{43\%} \\ 
    CoST & 0.784 & 0.624 & 0.540 & 0.494 & 0.428 & \textbf{26\%} \\ 
    TS2Vec & 0.655 & 0.632 & 0.599 & 0.577 & 0.517 & \textbf{31\%} \\ 
    \hline 
    
    \hline 
    
    \end{tabular}}
    \captionof{table}{Cross transfer learning MSE improvement (IMP) for self-supervised pre-training methods in various few-shot settings (10\%,25\%,50\%,75\%,100\%).}
    \captionsetup{justification=centering,singlelinecheck=false}
    \label{tab:simmtm_2}
    
  \end{minipage}%
  
\end{figure*}

\begin{figure*}[h]

\begin{minipage}{0.45\textwidth}

   \centering
    \setlength{\tabcolsep}{1.5pt}
    \scalebox{0.7}{
    \begin{tabular}{|c|c|c|c|c|} \hline 
     & \multicolumn{2}{|c|}{\makecell{Less PT Data \\ (250M samples)}} & \multicolumn{2}{|c|}{\makecell{More PT Data \\ (1Bsamples)}} \\ \hline

        \makecell{Data} & \makecell{w/o AP} & \makecell{w/ AP} & \makecell{w/o AP} & \makecell{w/ AP}\\ \hline 
        ETTH1 & 0.369 & \textbf{0.365} & 0.367 & \textbf{0.364} \\ 
        ETTH2 & \textbf{0.283} & 0.285 & \textbf{0.275} & {0.277} \\ 
        ETTM1 & 0.446 & \textbf{0.413} & 0.326 & \textbf{0.322} \\ 
        ETTM2 & 0.191 & \textbf{0.187} & 0.172 & \textbf{0.171} \\ 
        Weather & 0.159 & \textbf{0.154} & 0.161 & \textbf{0.158} \\ 
        Electricity & 0.179 & \textbf{0.169} & 0.174 & \textbf{0.166} \\ 
        Traffic & 0.521 & \textbf{0.518} & 0.522 & \textbf{0.514} \\ \hline
        \textbf{f-imp(\%)} & \multicolumn{2}{|c|}{\textbf{3\%}} & \multicolumn{2}{|c|}{\textbf{1.5\%}} \\
        
        \hline

        \end{tabular}
    }
    \captionof{table}{\scriptsize{Impact of Adaptive Patching(AP) in less pre-training (PT) and more pre-training (PT) data setting. Zero-shot results on FL 96 reported. AP generally improves the forecasting accuracy across both setups, but the impact is more when PT data is less as AP enables modelling at different resolutions in different layers of the model. [`w/': with, `w/o': `without'.]}}
    \captionsetup{justification=centering,singlelinecheck=false}
    \label{tab:abl_ap}
    
  \end{minipage}%
\hspace{0.5cm}
\begin{minipage}{0.45\textwidth}
   \centering
    \setlength{\tabcolsep}{1.5pt}
    \scalebox{0.7}{
    \begin{tabular}{|c|c|c|c|c|} \hline 
     & \multicolumn{2}{|c|}{\makecell{Less PT Data \\ (250M samples)}} & \multicolumn{2}{|c|}{\makecell{More PT Data \\ (1B samples)}} \\ \hline

        \makecell{Data} & \makecell{w/o RPT} & \makecell{w/ RPT} & \makecell{w/o RPT} & \makecell{w/ RPT}\\ \hline 

        ETTH1 & \textbf{0.365} & 0.36 & 0.366 & \textbf{0.364} \\ 
        ETTH2 & 0.285 & \textbf{0.28} & 0.285 & \textbf{0.277} \\ 
        ETTM1 & 0.413 & \textbf{0.384} & 0.341 & \textbf{0.322} \\ 
        ETTM2 & \textbf{0.187} & 0.194 & 0.18 & \textbf{0.171} \\ 
        Weather & \textbf{0.154} & 0.16 & \textbf{0.153} & 0.158 \\ 
        Electricity & \textbf{0.169} & 0.175 & 0.178 & \textbf{0.166} \\ 
        Traffic & \textbf{0.518} & \textbf{0.518} & 0.528 & \textbf{0.514} \\ \hline
        \textbf{f-imp(\%)} & \multicolumn{2}{|c|}{\textbf{0\%}} & \multicolumn{2}{|c|}{\textbf{3\%}} \\
        \hline

        \end{tabular}
    }
    \captionof{table}{\scriptsize{Impact of Resolution Prefix Tuning (RPT) in less pre-training (PT) and more pre-training (PT) data setting. Zero-shot results on FL 96 reported. RPT generally enhances the forecast performance especially when the volume and diversity in the pretraining (PT) data are high. 
In this setting, adding a learnable resolution prefix token greatly helps, as it enables the models to easily decouple the weights across resolutions. However, in less PT setup where the challenges in diversity modelling are not observed, RPT does not have much impact. [`w/': with, `w/o': `without'.]}}
    \captionsetup{justification=centering,singlelinecheck=false}
    \label{tab:abl_fpt}
  \end{minipage}%
\end{figure*}

\begin{figure*}[h]
\begin{minipage}{0.45\textwidth}

   \centering
    \setlength{\tabcolsep}{1.5pt}
    \scalebox{0.7}{

    \begin{tabular}{|c|c|c|}
        \hline

        & \multicolumn{2}{c|}{ \textbf{\ttmq~ $SL$ = 96: $FL$ = 24}} \\ \hline
        
        & \textbf{w/o RPT} & \makecell{\textbf{w/ RPT}}  \\ \hline
        ETTH1 & 0.373 & \textbf{0.358}  \\ 
        ETTH2 & 0.180 & \textbf{0.179}  \\ 
        ETTM1 & 0.559 & \textbf{0.387} \\ 
        ETTM2 & 0.127 & \textbf{0.108}  \\ 
        Weather & \textbf{0.103} & \textbf{0.103}  \\ 
        Electricity & 0.208 & \textbf{0.201}  \\ 
        Traffic & 0.754 & \textbf{0.740}  \\ 
        \hline 
        \makecell{\textbf{IMP (\%)}} & \multicolumn{2}{c|}{\makecell{\textbf{ 8\%}}}  \\ 
        \hline 
        \end{tabular}
        
    }
    \captionof{table}{\scriptsize{Impact of RPT in less context setting ($SL$ = 96). Zero-shot results on FL 24 reported. RPT helps in scenarios when the context length ($sl$) is short. In these scenarios, automatically detecting the resolution becomes a challenge for the model. Hence, by explicitly fusing the resolution information as a prefix, we can enhance the model's ability to learn effectively across resolutions. [`w/': with, `w/o': `without'.]}}
    \captionsetup{justification=centering,singlelinecheck=false}
    \label{tab:abl_rpt_2}
  \end{minipage}%
\end{figure*}

\section{Model Insights and Explanation}
\label{appendix:explain}
\subsection{Dataset preparation for TTM embedding analysis}
\label{app:sec:Dataset preparation for TTM embedding analysis}
\begin{figure*}
\centering 
\includegraphics[width=0.95\textwidth]{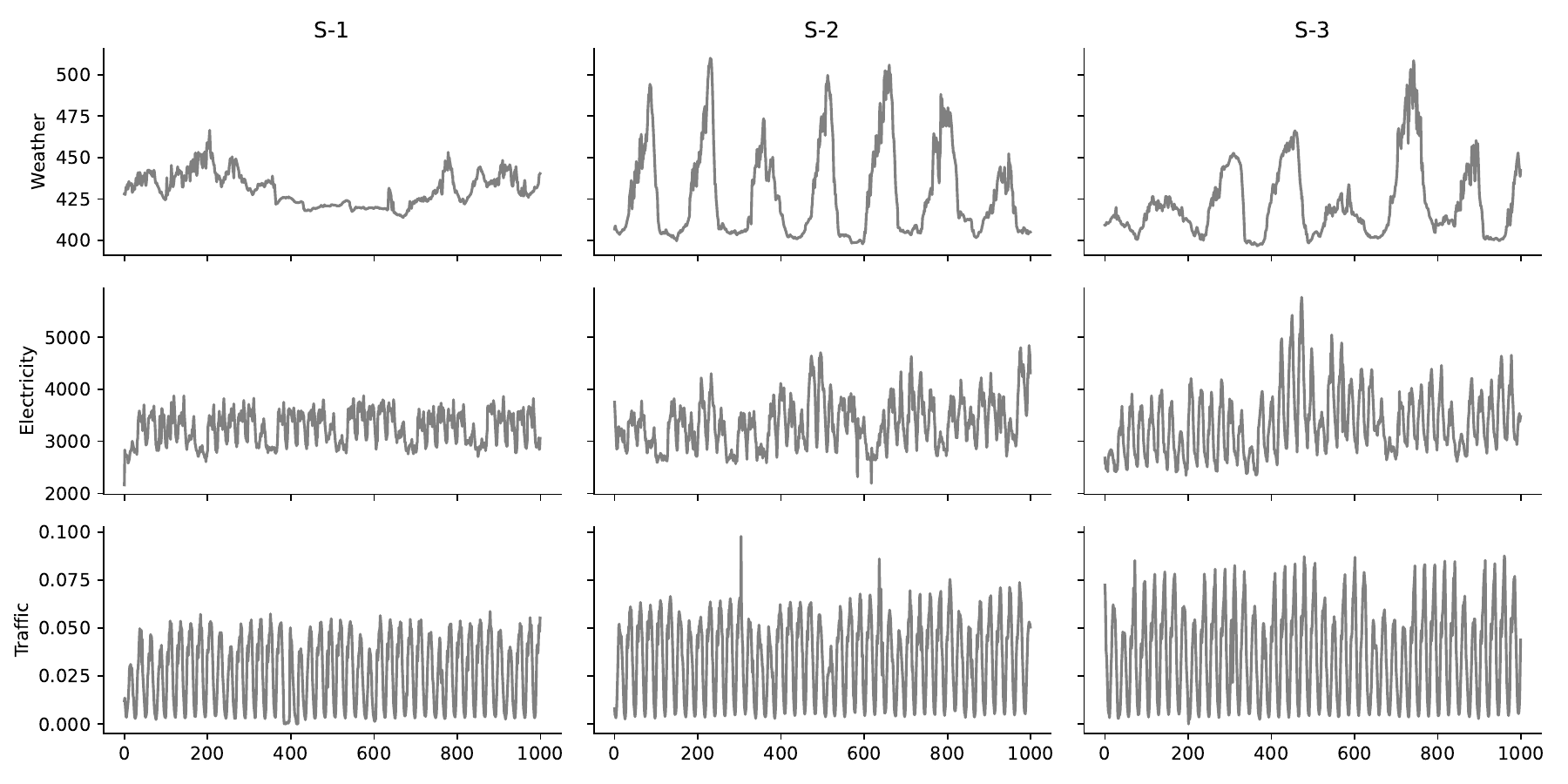}
\caption{Data segments for TTM embedding analysis.}
\label{fig:data-segments}
\end{figure*}

To understand the representation obtained from TTM encoder, we have carried out a controlled analysis, using 3 datasets with varying observation frequency, viz., (1) weather (10 min), (2) electricity (1 hour), and (3) traffic (1 hour). We have selected three temporally distinct non-overlapping windows of length 1024 from each dataset (Figure~\ref{fig:data-segments}). The selection criteria of these segments are distinct mean and standard deviation measures. From each of these segment, 512 context length windows are extracted in a rolling window fashion. Embedding vector from the encoder is collected at backbone output. Each of these representation tensors are flattened, and Principal Component Analysis~(PCA) is carried out on the whole dataset. The project on the first two principle components is used to obtain the figure~\ref{fig:ttm-embedding}.

\subsection{Channel Attention Map}
The channel mixing block in the decoder of TTM consists of a gated attention block that produces an attention weight for each feature across channels. We have considered the mean attention weight across features and data samples to derive the feature contribution. We have used the model finetuned on the Bikesharing dataset for this purpose. Bikesharing data includes exogenous variables, viz. temperature, humidity, wind speed, etc. We analyzed the mean attention of the model across these exogenous variables for the forecast of rental bike count (Figure~\ref{fig:bikesharing-exog-explain}). As we observe, these attention weights highly correlate with the general data characteristics of his data, wherein - bike rentals are highly influenced by weather and holiday signals. Thus, TTM fine-tuning process is quick as well as explainable.

\section{Limitations and Future Work}
\label{appendix:limitations}
TTM is currently focused solely on forecasting tasks, similar to other forecast pretraining models like Moirai~\cite{moirai}, Chronos~\cite{chronos}, and TimesFM~\cite{timesfm}. However, recent models such as Moment and GPT4TS are taking initial steps to expand their capabilities across multiple downstream tasks, including classification, regression, and anomaly detection. Inspired by these advancements, we plan to extend TTM's functionality to encompass a broader range of downstream tasks.

Another limitation of TTM is the need to train different models for different context length settings. Due to its non-transformer-based architecture, TTM is sensitive to context lengths. Consequently, in this paper, we introduce three variants of TTM, each optimized for different context length settings. Looking ahead, we aim to enhance TTM's backbone to automatically adapt to dynamically varying context lengths. 

In addition, existing pre-trained models like lag-llama, Moirai support probabilistic forecasting while TTM currently supports only point forecasting. We plan to extend TTM with distribution heads to facilitate probabilistic forecasts in future work.




\end{document}